%% file: main.tex
\documentclass{article}

\PassOptionsToPackage{numbers, compress}{natbib}
\usepackage[preprint]{neurips_2025}

\usepackage{microtype}
\usepackage{graphicx}
\usepackage{subfigure}
\usepackage{booktabs} 

\input{math_commands.tex}

\usepackage{url}
\usepackage[table]{xcolor}
\usepackage{graphicx}
\usepackage{booktabs}
\usepackage{multirow}
\usepackage{makecell}
\usepackage{enumitem}

\usepackage{amsmath}
\usepackage{amssymb}
\usepackage{mathtools}
\usepackage{amsthm}

\usepackage{wrapfig}

\usepackage{tablefootnote}

\usepackage{colortbl}

\usepackage[colorlinks = true,
            linkcolor = blue,
            urlcolor  = blue,
            citecolor = blue,
            anchorcolor = blue]{hyperref}

\usepackage{pifont}

\usepackage[capitalize,noabbrev]{cleveref}

\usepackage{siunitx}

\newcommand{\xhdr}[1]{\textbf{#1}.}

\usepackage{xspace}
\newcommand{\name}{FORMED\xspace}

\theoremstyle{plain}
\newtheorem{theorem}{Theorem}[section]

\theoremstyle{definition}
\newtheorem{definition}[theorem]{Definition}

\theoremstyle{remark}

\usepackage[textsize=tiny]{todonotes}

\newcommand{\std}[1]{\scriptsize{$\pm$#1}}
\newcommand{\ie}{\textit{i.e.}}
\newcommand{\eg}{\textit{e.g.}}

\title{Repurposing Foundation Model for Generalizable Medical Time Series Classification}

\author{
  Nan Huang\\
  University of North Carolina at Charlotte\\
  NC, US \\
  \texttt{nhuang5@uncc.edu} \\
  \And
  Haishuai Wang \\
  Zhejiang University\\
  Zhejiang, China \\
  \texttt{haishuai.wang@zju.edu.cn} \\
  \And
  Zihuai He \\
  Stanford University\\
  Stanford, CA, US \\
  \texttt{zihuai@stanford.edu} \\
  \And
  Marinka Zitnik \\
  Harvard University\\
  Boston, MA, US \\
  \texttt{marinka@hms.harvard.edu} \\
  \And
  Xiang Zhang \\
  University of North Carolina at Charlotte\\
  NC, US \\
  \texttt{xiang.zhang@charlotte.edu} \\
}

\begin{document}

\maketitle

\begin{abstract}
  \input{sections/01-abstract}
\end{abstract}

\input{sections/02-introduction}

\input{sections/03-method}

\input{sections/04-experiment}

\input{sections/05-conclusion}

\newpage
\bibliography{references}
\bibliographystyle{unsrtnat}

\clearpage
\appendix
\onecolumn
\input{sections/99-appendix}


\input{sections/100-checklist}

\end{document}

%% file: math_commands.tex
\usepackage{amsmath,amsfonts,bm}

\def\eqref#1{equation~\ref{#1}}

\def\ceil#1{\left\lceil #1 \right\rceil}

\def\1{\bm{1}}

\def\vd{{\bm{d}}}

\def\vm{{\bm{m}}}

\def\vx{{\bm{x}}}
\def\vy{{\bm{y}}}

\def\evm{{m}}

\def\mE{{\bm{E}}}

\def\mH{{\bm{H}}}

\def\mM{{\bm{M}}}

\def\mQ{{\bm{Q}}}

\def\mX{{\bm{X}}}

\def\mZ{{\bm{Z}}}

\DeclareMathAlphabet{\mathsfit}{\encodingdefault}{\sfdefault}{m}{sl}
\SetMathAlphabet{\mathsfit}{bold}{\encodingdefault}{\sfdefault}{bx}{n}
\newcommand{\tens}[1]{\bm{\mathsfit{#1}}}

\def\tE{{\tens{E}}}

\def\tH{{\tens{H}}}

\def\tQ{{\tens{Q}}}

\newcommand{\E}{\mathbb{E}}

\newcommand{\R}{\mathbb{R}}

\DeclareMathOperator*{\argmin}{arg\,min}

%% file: sections/01-abstract.tex
Medical time series (MedTS) classification suffers from poor generalizability
in real-world deployment due to inter- and intra-dataset heterogeneity, such as varying
numbers of channels, signal lengths, task definitions, and patient characteristics.
To address this, we propose \name, a novel framework for repurposing a backbone foundation model, pre-trained on generic time series, to enable \textbf{highly generalizable} MedTS classification on unseen datasets.
\name combines the backbone with a novel classifier comprising two components: (1) task-specific channel embeddings and label queries, dynamically sized to \textbf{match any number of channels and target classes}, and (2) a shared decoding attention layer, jointly trained across datasets to \textbf{capture medical domain knowledge} through task-agnostic feature-query interactions. After repurposing, \name achieves seamless adaptation to unseen MedTS datasets through lightweight label query training (0.1\% of parameters), eliminating the need for full fine-tuning or architectural redesign.
We evaluate \name on 5 diverse MedTS datasets, benchmarking against 11 Task-Specific Models (TSM) and 4 Task-Specific Adaptation (TSA) methods. Our results demonstrate \name's dominant performance, achieving up to \textbf{35\% absolute} improvement in \textbf{F1}-score (on ADFTD dataset) over specialized baselines.
Further analysis reveals consistent generalization across varying channel configurations, time series lengths, and clinical tasks, which are key challenges in real-world deployment.
By decoupling domain-invariant representation learning from task-specific adaptation, \name establishes a scalable and resource-efficient paradigm for foundation model repurposing in healthcare. This approach prioritizes clinical adaptability over rigid task-centric design, offering a practical pathway for real-world implementation.
\textit{Code release upon acceptance.}

%% file: sections/02-introduction.tex
\section{Introduction}

\begin{figure*}[htb]
    \centering
    \includegraphics[width=0.95\textwidth, trim={0 0.225cm 0 0.25cm}, clip]{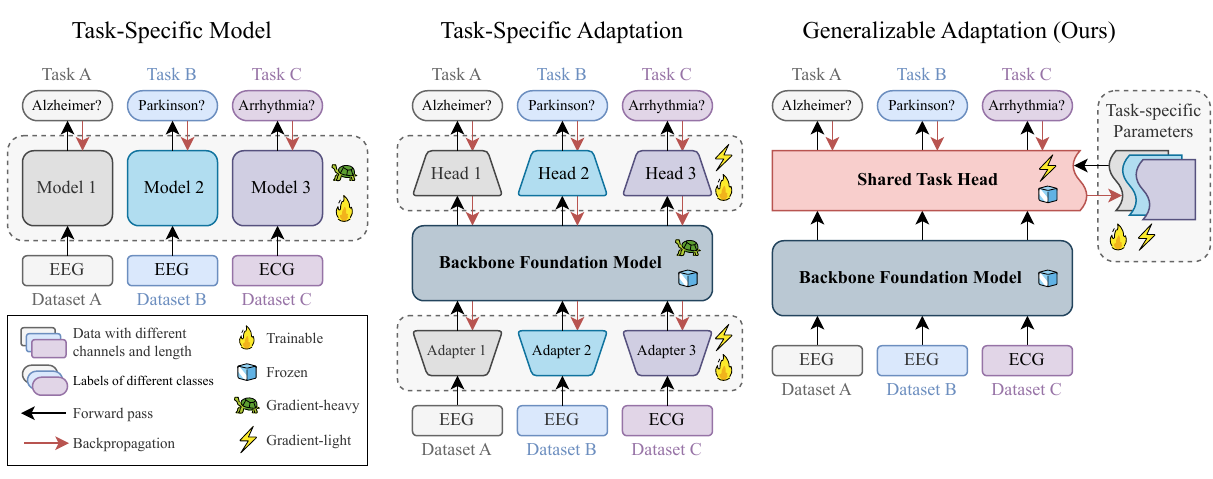}
    \caption{Paradigms of building models for different MedTS classification tasks. \textbf{Task-Specific Model (TSM):} Traditional classification models are designed for specific input shape and output classes, thus require retraining from scratch for each new dataset. \textbf{Task-Specific Adaptation (TSA):} By using a pre-trained and fixed backbone foundation models, the adaptation to new datasets requires training fewer parameters for each dataset, such as pre- and post-backbone adapters, which makes the combined model no longer applicable to other tasks, lacking generalization across tasks, and more prone to overfitting. \textbf{Generalizable Adaptation (GA):} Generalizable adaptation is a post-backbone adaptation module that is shared across tasks of different datasets, which carries domain knowledge and transferable to unseen datasets with training of lightweight task-specific parameters, offering both generalizability and robustness against overfitting.
    }
    \label{fig:ofo-vs-ofa}
    \vskip -0.2in
\end{figure*}

%

Medical time series (MedTS) classification, such as on electroencephalograms (EEG) and electrocardiograms (ECG),
is critical for diagnosing a wide spectrum of medical conditions, including Alzheimer's Disease (AD; \citet{EEG-AD}), Parkinson's Disease (PD; \citet{EEG-PD-1,EEG-PD-2}), and heart Arrhythmia \citep{ECG-AR-2}.
Despite significant advancements in developing deep learning models for these tasks, their effective generalization across diverse datasets, sometimes even among individual patients within the same dataset, remains a significant hurdle. This limitation critically obstructs the translation of advanced predictive algorithms into reliable real-world clinical applications.

Several unique challenges inherent to MedTS data compound the poor generalizability. Firstly, \textbf{inter-dataset heterogeneity} arises from variations in physiological data domains, data collection equipment, and experimental protocols, leading to differences in the \textit{number of channels}, \textit{sample durations}, \textit{sampling rates}, and \textit{diagnostic targets} \citep{Survey-DL-in-MedTS}. Secondly, \textbf{intra-dataset heterogeneity} persists even within single datasets, with variations occurring across recording times, experimental sessions, and, most importantly, \textit{among patients due to unique physiological characteristics} \citep{Medformer,Survey-DL-in-MedTS}. This often leads to models overfitting to training data and failing to generalize to unseen patients. Thirdly, \textbf{data insufficiency} is a persistent issue: the high cost of data collection and privacy concerns often result in small MedTS datasets \citep{Survey-challenges-in-MedTS}, making it difficult to train robust models capable of addressing the aforementioned heterogeneity \citep{Survey-DL-in-MedTS}.

Previous attempts to tackle these issues, such as employing Task-Specific Adaptation (TSA; see \cref{fig:ofo-vs-ofa}) in models like \citet{BIOT}, have shown limited success. These methods may unintentionally focus on extracting features relevant only to the initial training task, thereby failing to generalize effectively to new datasets or different medical conditions, as evidenced by marginal or even negative performance gains through pre-training. While on the other hand, recent advancements in foundation models for time series offer a promising avenue. Despite their \textit{predominant focus on forecasting tasks} \citep{Survey-TS-LLM,Survey-Transformer-TS}, they demonstrate the ability to learn generic representations of time series data \citep{Survey-TSA-FM}, thanks to pre-training on large-scale general time series data. This can be beneficial for MedTS classification tasks as well.
Our pilot study indicates that directly adapting these models for MedTS classification is better than TSA models trained from scratch, but still fall short in capturing the intricate patterns necessary for specific diagnostic tasks when compared to state-of-the-art (SOTA) Task-Specific Models (TSMs; see \cref{fig:ofo-vs-ofa}). This is primarily due to their lack of ability to capture the task-agnostic domain knowledge, which is crucial for generalization across datasets.


To address these limitations, this paper introduces \name (\textbf{Fo}undation model \textbf{R}epurposed for \textbf{Med}ical time series classification), a novel approach designed to repurpose foundation models for MedTS classification. \name aims to achieve \textbf{Generalizable Adaptation} (GA; see \cref{fig:ofo-vs-ofa}). This is achieved by utilizing a pre-trained foundation model as its backbone to capture \textbf{generic temporal features}, and integrating a novel classifier design to handle MedTS heterogeneity, by architecturally separating \textbf{medical domain knowledge} from \textbf{task-specific knowledge}. This allows the model to effectively learn and utilize both task-agnostic and task-specific features, enabling seamless handling of datasets with arbitrary channel configurations, dynamic time series lengths, and diverse diagnostic targets across multiple tasks. Therefore, \name enables the backbone model to effectively leverage the commonalities across datasets, while also being flexible enough to adapt to the unique characteristics of each dataset, thus achieving generalization across datasets and tasks.

To facilitate this research, we curate a comprehensive \textit{repurposing cohort} comprising five MedTS datasets (two on ECG and three on EEG), totaling approximately 340,000 samples or 90 million time-points.
These datasets exhibit diverse channel configurations (ranging from 12 to 33 channels), varied diagnostic tasks (from binary neurological to 5-class cardiovascular classification), and differing dataset sizes. This provides different levels of difficulties (both inter- and intra-dataset) for the model to learn from, and serves as a robust training and evaluation platform for the proposed method.



\name is strongly supported by empirical results on two aspects: First, for datasets partially included in the repurposing cohort, \name achieves state-of-the-art level performance on unseen patients, outperforming 15 TSA and TSM models across all datasets. Second, for completely new datasets not included in the cohort, \name can be efficiently adapted by updating only a small proportion of parameters while outperforming the baseline TSA model, and shows robust adapt-time scaling performance with the amount of trainable parameters. This demonstrates the model's ability to generalize across datasets and tasks, and its potential for real-world applications in healthcare.

\section{Related Work}
\label{sec:related-work}

\xhdr{Foundation Models for General Time Series}
To date, no foundation model has been specifically designed for time series classification tasks, let alone MedTS classification; instead, recent advances in time series foundation models mainly concentrate on forecasting tasks \citep{Survey-TSA-FM}.
Given their success in forecasting, re-purposing these models for MedTS classification is a tempting prospect, yet it presents significant theoretical and practical challenges. These models often have major limitations, such as an inherent design for \textit{univariate time series} in a channel-independent fashion \citep{PatchTST}, and requiring \textit{TSAs} that prevent them from being directly applicable to MedTS classification tasks \citep{TEMPO,TEST,LLM4TS}. Given that medical time series are typically multi-variate, a critical aspect of our repurposing framework is to effectively integrate the information from multi-channel features extracted by these backbones.

For instance, models like Time-LLM \citep{Time-LLM}, UniTime \citep{UniTime} and GPT4TS \citep{GPT4TS} are backboned on large language models. Consequently, they naturally handle time series data in a univariate manner, lacking the ability to integrate information across multiple channels crucial for MedTS classification.
Moreover, consistent with the findings of \citet{LLM4TS-Not-Useful}, our empirical observations show that LLM-based time series foundation models do not always achieve optimal performance, even on general time series datasets.
Similarly, while TimeGPT \citep{TimeGPT} and TimesFM \citep{TimesFM} are pre-trained on large scale time series data, they typically operate under a channel-independence assumption, treating co-evolving multivariate time series data as a collection of independent univariate series. This shares the same limitation for direct application to multi-channel MedTS. Our proposed repurposing framework is specifically designed to address this by incorporating mechanisms to accommodate the multi-channel nature of MedTS data, allowing for effective integration of information across channels.

UniTS \citep{UniTS} stands out as capable of handling multivariate time series data and has been trained on multiple task domains including classification. However, its scale and design often necessitate fine-tuning the entire model or employing prompt learning for optimal performance. This approach is both computationally expensive (see \cref{fig:ofo-vs-ofa}) and data-greedy due to the vast number of parameters to tune, rendering it less suitable for often small-scale MedTS datasets.


Despite their efforts and successes, current foundation models require significant adaptations. Thus, \textbf{a key challenge, which \name directly addresses, is the adaptation of these powerful but often channel-independent or forecasting-focused models to the multichannel classification demands of MedTS}. This is achieved by integrating dedicated architectural components to address the complexities and multi-channel nature of MedTS. While all the aforementioned models represent potential backbones for our repurposing framework, due to resource limitations and the primary goal to validate the efficacy of our proposed framework, we selected the advanced TimesFM \citep{TimesFM} as our backbone model for this study for its outstanding zero-shot forecasting performance.


\xhdr{Adaptation of Foundation Models for MedTS Classification}

General-purpose foundation models typically require specific techniques to be effectively adapted for downstream tasks. Common approaches include \textit{prompting}, \textit{fine-tuning}, \textit{re-programming}, and our proposed \textbf{re-purposing}.
Although a detailed comparison of all techniques is beyond the immediate scope (further discussion in \cref{sec:adaptation-comparison}), we will focus here on the distinction between re-programming and re-purposing, as this differentiation is central to our proposed approach for MedTS classification.


\textit{Re-programming} often involves reusing a pre-trained model's backbone (\eg, its Transformer layers) without altering its internal weights, but wrapping it with new input adapters and task-specific output heads (as illustrated by the TSA approach in \cref{fig:ofo-vs-ofa}). While this can adapt a model to a new data domain or task type, a significant drawback is that the resulting model often \textbf{loses its general-purpose nature}. Both the input adapters and task heads become highly specialized, and \textbf{cannot be reused} in future datasets with different configurations, hindering the model's ability to generalize across different tasks, datasets, or to handle the inherent heterogeneity within MedTS \citep{LLM4TS-Not-Useful}.

\textit{Re-purposing}, as introduced in this work with our \name framework, takes a different philosophical approach. It aims to adapt a pre-trained foundation model (often one excelling in tasks like forecasting) to a new class of tasks---in our case, MedTS classification. This is achieved with thoughtful modifications, particularly in how task-specific knowledge is integrated, while striving to maintain the model's core learned representations. The goal is for the repurposed model to serve as a robust and generalizable tool within the target domain (MedTS classification), capable of being efficiently applied to new datasets and diagnostic challenges. Its emphasis on generalizability, data-efficiency, and leveraging domain insights makes re-purposing particularly suitable for adapting powerful, channel-independent time series foundation models to the complexities of multi-channel MedTS classification, and creating a new foundation model for the field.



\xhdr{Forecasting versus Classification in Time Series}
The fundamental differences between time series forecasting and classification are key to understanding the challenges in adapting existing foundation models. \textbf{Forecasting} typically involves predicting future sequence values within the same domain as the input (a sequence $\rightarrow$ sequence mapping) \citep{Survey-TS-forecasting,EEG-Transformer-forecasting}, often by extrapolating patterns \textbf{within individual channels}. In contrast, \textbf{classification} maps an input sequence to a distinct categorical label (a sequence $\rightarrow$ category mapping) \citep{Survey-TS-classification}. This frequently requires synthesizing complex patterns across \textbf{multiple interacting channels} to derive a diagnostic outcome (\eg, diagnosing disease from multi-channel EEG; \citep{EEG-Transformer-classification})---a process inherently different from predicting future signal values. This intrinsic divergence in objectives and the nature of data interpretation means that adapting a forecasting foundation model for MedTS classification is \textbf{NOT a simple modification of the output layer}. It demands a more comprehensive re-purposing strategy, such as our \name framework, designed to bridge these task-specific requirements and effectively handle the complexities of multi-variate medical data.



%% file: sections/03-method.tex
\section{Problem Statement}
\label{sec:problem}

\begin{figure*}[htb]
    \centering
    \includegraphics[width=0.95\textwidth, trim={0.25cm 0 1.25cm 0.25cm}, clip]{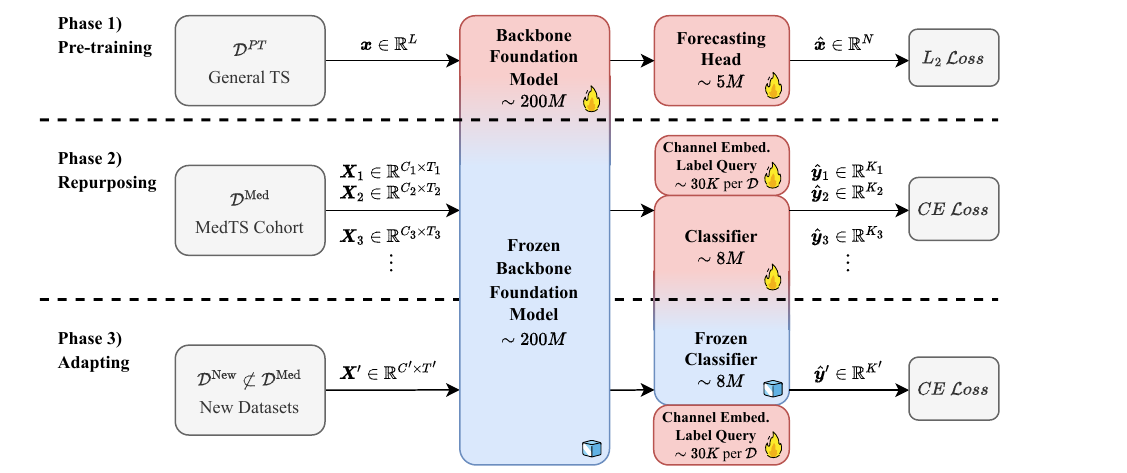}
    \caption{
        The three-stage process of adapting a time series foundation model for MedTS classification tasks. 1) \textbf{Pre-training} is already done on diverse general time series datasets with forecasting tasks. 2) \textbf{Repurposing} the foundation model involves changing the forecasting head to a classification head, while keeping the rest of the model fixed, and the new model is then trained on a cohort of MedTS datasets to capture domain knowledge in MedTS. 3) \textbf{Adapting} the repurposed model to the new MedTS datasets, where only the minimal task-specific parameters are trained, leveraging the previously learned domain knowledge from the repurposed model.
    }
    \label{fig:three-stage}
\end{figure*}

Foundation models, pre-trained on diverse forecasting tasks, have demonstrated a strong capability to capture general time series patterns. However, transforming these into general-purpose classification models, especially for MedTS, is non-trivial. This section formally defines the problem and the core concepts underpinning our proposed two-stage adaptation process: repurposing and adapting.


\begin{definition}
    \textbf{Repurposing}: The process of changing the objective of a pre-trained foundation model to a new class of tasks for which it was not originally trained. This involves introducing and training a relatively small, adaptable output network while keeping the pre-trained backbone fixed.
    \label{def:repurposing}
\end{definition}

Let the original pre-trained model consists of a backbone $f: \R^T \rightarrow \R^{L\times D}$ for extracting features from a univariate time series of length $T$ into $L$ patched tokens of dimension $D$,
and an original task head (\eg, for forecasting, $g: \R^{L\times D} \rightarrow \R^{N}$). We leverage the frozen backbone $f$ for representation learning. For multivariate MedTS input $\mX\in\R^{C \times T}$ with $C$ channels, the backbone is applied channel-wise to extract features:
\begin{equation}
    \begin{aligned}
        \mathbf{f}: \R^{C \times T} \rightarrow \R^{C\times L\times D} \\
        \Leftrightarrow \mathbf{f}(\mX) := \left[f(\mX_{c,:})\right]_{c=1}^C
    \end{aligned}
    \label{eq:channel-wise-feature}
\end{equation}

We then introduce a novel, trainable classification head $h_\theta$. This head is designed to be adaptable to specific task characteristics, such as the number of input channels $C$ and the number of output classes $K$, through learnable task-specific parameters: \textbf{Channel Embedding} $\mE\in\R^{C\times D}$ and \textbf{Label Queries} $\mQ\in\R^{K\times D}$. The mapping becomes:
\begin{equation}
    \begin{aligned}
        \left.h_{\theta}\right|_{\mQ,\mE}:                                       & \R^{C\times L\times D} \rightarrow \Delta^K \\
        \Rightarrow \left.\left(h_\theta\circ\mathbf{f}\right)\right|_{\mQ,\mE}: & \R^{C\times T} \rightarrow \Delta^K
    \end{aligned}
\end{equation}
where $\Delta^K = \left\{\vd\in[0,1]^K: \sum_{i=1}^{K}{d_i}=1\right\}$ is the probability simplex for $K$ classes.

During the \textbf{repurposing stage}, $h_\theta$ containing shared parameter $\theta$ along with collections of task-specific embeddings $\tE=\left\{\mE_i\right\}$ and $\tQ=\left\{\mQ_i\right\}$ are trained across a cohort of diverse MedTS datasets $\mathcal{D}^{\textrm{Med}}=\left\{\mathcal{D}^{\textrm{Med}}_i\right\}$. The objective is to learn domain knowledge for MedTS classification within $\theta$. This translates to minimizing the cross-entropy loss $\mathcal{L}_{\text{CE}}$:
\begin{equation}
    \theta^\ast,\tE^\ast,\tQ^\ast = \argmin_{\theta,\tE,\tQ} \E_{i,(\mX_i,\vy_i)\in\mathcal{D}_i^{\textrm{Med}}}\left[\mathcal{L}_{\text{CE}}\left(\left.h_\theta\right|_{\mQ_i,\mE_i}\left(\mathbf{f}(\mX_i)\right), \vy_i\right)\right]
    \label{eq:repurposing-objective}
\end{equation}

\begin{definition}
    \textbf{Adapting}:
    The process of applying the repurposed model (with fixed $\theta^\ast$ and frozen $\mathbf{f}$) to new, unseen MedTS datasets or tasks. This involves learning only a minimal set of new task-specific parameters (new $\mE'$ and $\mQ'$) for the new dataset.
    \label{def:adapting}
\end{definition}

For a new dataset $\mathcal{D}^{\textrm{New}}$ (with potentially different $C'$ channels, $T'$ time steps, and $K'$ classes), the pre-trained backbone $\mathbf{f}$ and the shared parameters $\theta^\ast$ of the classifier $h_{\theta^\ast}$ remain frozen. Only new Channel Embeddings $\mE'\in\R^{C'\times D}$ and Label Queries $\mQ'\in\R^{K'\times D}$ are initialized and trained:
\begin{equation}
    \begin{aligned}
        {\mE'}^\ast,{\mQ'}^\ast = & \argmin_{\mE',\mQ'} \E_{(\mX',\vy')\in\mathcal{D}^{\textrm{New}}}\left[\mathcal{L}\left(\left.h_{\theta^\ast}\right|_{\mQ',\mE'}\left(\mathbf{f}(\mX')\right), \vy'\right)\right] \\
    \end{aligned}
    \label{eq:adapting-objective}
\end{equation}

This two-stage process (illustrated in Figure \ref{fig:three-stage}) allows the model to learn general MedTS domain knowledge and efficiently specialize to new tasks with minimal new data and computation.

\section{Model Architecture}
\label{sec:model}

Our \name framework repurposes a pre-trained time series foundation model to serve as a versatile MedTS classification tool. It comprises two main parts: a frozen backbone feature extractor and our novel attention-based classifier designed for adaptability and generalizability.

\subsection{Backbone Feature Extractor for Variable Length Time Series}

\begin{figure*}[htb]
    \centering
    \includegraphics[width=\textwidth, trim={0.2cm 0 0.8cm 0}, clip]{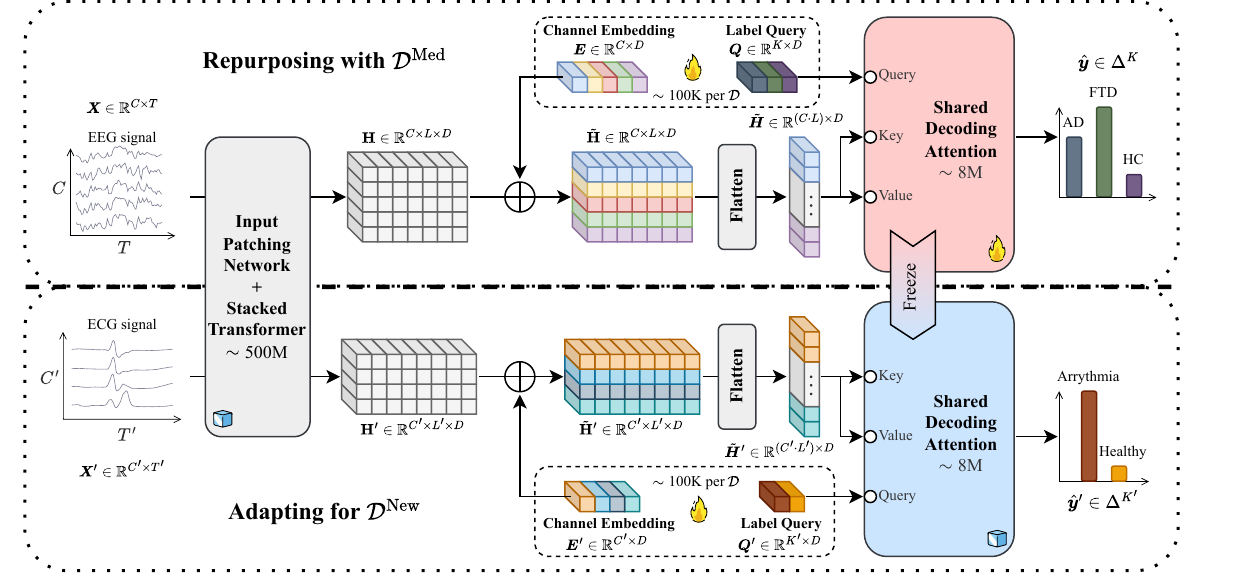}
    \vskip -0.1in
    \caption{The architecture of the proposed model in repurposing and adapting.
    The backbone foundation model acts as a feature extractor and remains frozen all the time. The \textbf{Channel Embeddings} (CEs) and \textbf{Label Queries} (LQs) are task-specific parameters that are learned during both repurposing and adapting, and new ones will be created and learned if encountering new datasets. The \textbf{Shared Decoding Attention} (SDA) is a shared Transformer decoder layer that captures the interaction between all the features and classes, which once get trained on curated MedTS datasets $\mathcal{D}^{\text{Med}}$ during repurposing, will be fixed and reused when adapting to all future datasets and tasks $\mathcal{D}^{\text{New}}$. The $\oplus$ denotes broadcast addition.
    }
    \label{fig:architecture}
    \vspace{-3mm}
\end{figure*}

We employ a powerful pre-trained time series foundation model as the backbone feature extractor. For this work, we selected TimesFM \citep{TimesFM} due to its demonstrated ability to capture complex temporal patterns from variable-length time series, pre-trained on a vast and diverse corpus of general time series data. The backbone's primary role is to transform an input univariate time series $\vx\in\R^T$ into a sequence of rich feature tokens $\mH\in\R^{L\times D}$. As defined in \cref{eq:channel-wise-feature}, for multi-channel MedTS data $\mX \in \R^{C \times T}$, this backbone is applied independently to each channel, yielding stacked feature tokens $\tH \in \R^{C \times L \times D}$. The internal architecture of TimesFM (\eg, its patching mechanism and Transformer layers) is kept frozen throughout our framework and detailed in \cref{sec:implementation}. The crucial output for our purpose is $\tH$, which serves as the input to our novel classifier.

\subsection{Attention-based Classifier for Varying Channel and Class}

Traditional approaches often use a simple linear layer or convolution layer for classification \citep{TST,BIOT}, which can be restrictive when dealing with the inherent variability of MedTS data (\eg, varying numbers of channels, classes, and sequence lengths). To address these unique challenges, we propose a novel attention-based classifier built upon a Transformer decoder layer \citep{Transformer}, inspired by advancements in vision tasks \citep{DETR, CV-Attention-Classification} but with key modifications tailored for MedTS. This classifier comprises three main components: Channel Embeddings (CEs), Label Queries (LQs), and a Shared Decoding Attention (SDA) mechanism.

\xhdr{Channel Embeddings (CEs)} MedTS are inherently multi-variate. To enable the model to distinguish information from different physiological channels and understand their task-specific relevance, we introduce learnable Channel Embeddings $\mE\in\R^{C\times D}$. For a given dataset with $C$ channels, these embeddings are broadcast-added to the corresponding channel-wise feature tokens $\tH$ (from \cref{eq:channel-wise-feature}) to produce "channel-aware" feature tokens $\tilde{\tH} \in \R^{C\times L \times D}$:
\begin{equation}
    \tilde{\tH}{c,l,:}=\tH{c,l,:}\oplus\mE_{c,:}\quad\forall l\in\left\{1,2,\cdots,L\right\}, c\in\left\{1,2,\cdots,C\right\}
    \label{eq:channel-embeddings}
\end{equation}
These CEs are task-specific and learned during both repurposing (part of $\tE$ in \cref{eq:repurposing-objective}) and adapting ($\mE'$ in \cref{eq:adapting-objective}).

\xhdr{Label Queries (LQs)} To handle varying numbers of diagnostic classes ($K$) per task and provide distinct learnable "anchors" for each class, we use Label Queries $\mQ\in\R^{K\times D}$. Each row $\mQ_{i,:}$ is a learnable embedding representing the $i$-th class. These queries actively seek evidence for their respective classes within the channel-aware feature tokens. Like CEs, LQs are task-specific and learned during repurposing and adapting. In practice, we employ $k$ learnable queries for each class to capture potentially complex or multiple defining patterns, where $k$ is a hyperparameter determining the number of distinct "perspectives" or "sub-pattern detectors" for each class. This results in a total of $K\cdot k$ queries in $\mQ\in\R^{(K\cdot k)\times D}$. Each group of $k$ queries corresponding to a specific class independently attends to the feature tokens to gather evidence.

\xhdr{Shared Decoding Attention (SDA)} The core of our classifier is the SDA mechanism, a single Transformer decoder layer whose parameters ($\theta$ in \cref{eq:repurposing-objective}) are shared across all datasets in the repurposing cohort and then frozen during adaptation to new tasks. The SDA takes all $K\cdot k$ Label Queries from $\mQ$ as queries, and the flattened, channel-aware feature tokens $\texttt{Flatten}(\tilde{\tH}) \in \R^{(C\cdot L)\times D}$ as keys and values. It performs multi-head attention followed by an \texttt{FeedForwardNetwork} to produce an initial set of logits $\hat{\vy}_{\text{raw}} \in \R^{K \cdot k}$:
\begin{equation}
    \hat{\vy}_{\text{raw}} = \texttt{FeedForwardNetwork}\left(\texttt{MultiHeadAttention}(\mQ, \texttt{Flatten}(\tilde{\tH}), \texttt{Flatten}(\tilde{\tH}))\right)
    \label{eq:sda-mechanism-raw-logits}
\end{equation}
Since we have $k$ queries (and thus $k$ raw logits) for each of the $K$ classes, these $k$ logits are then averaged to produce a single, final logit for each class, resulting in the final class logits $\hat{\vy} \in \R^K$:
\begin{equation}
    \hat{\vy}_j = \frac{1}{k} \sum_{i=1}^{k} (\hat{\vy}_{\text{raw}})_{(j-1)k + i} \quad \forall j\in\left\{1,2,\cdots,K\right\}
    \label{eq:average_k_logits}
\end{equation}
These final logits $\hat{\vy}$ are then used with a \texttt{softmax} function for probability prediction and loss calculation. Critically, the parameters within the \texttt{MultiHeadAttention} and the \texttt{FeedForwardNetwork} (collectively $\theta$) are independent of the specific number of input channels $C$, token length $L$, or the total number of queries $K\cdot k$. This architectural choice forces SDA to learn generalizable interaction patterns while allowing for richer, more nuanced class representations.

%% file: sections/04-experiment.tex
\section{Experiments}
\label{sec:experiment}

Here we describe the experimental setup, including the datasets chosen to evaluate \name against key MedTS challenges, the baselines selected to highlight the advantages of our repurposing framework, and evaluation metrics. Additional training details are included in \cref{sec:experiment-setup}.

\xhdr{Datasets}
To thoroughly evaluate \name, we curated a \textbf{MedTS cohort} of 5 diverse datasets for the crucial repurposing stage (\cref{fig:three-stage}).  This cohort, comprising two ECG datasets and three EEG datasets (details in \cref{tab:dataset}), provides a breadth of configurations and task types. This enables the \textbf{learning of robust, generalizable MedTS domain knowledge} during repurposing. Critically, these datasets span a variety of combinations of channels, sampling rates, sample durations, diagnostic labels, and overall sizes. This inherent diversity is intentional, enabling us to directly \textbf{assess \name's ability to handle inter-dataset heterogeneity} during repurposing.

Furthermore, all datasets within the cohort and for subsequent adaptation are split following strict patient-independent settings \citep{Medformer,PatientIndependent}. This ensures that the test sets contain subjects entirely unseen during training, rigorously \textbf{evaluating the model's robustness against intra-dataset heterogeneity} and its capacity to generate to new patients rather than memorizing subject-specific features.

To specifically test the adapting stage and \name's generalization to entirely new, unseen tasks and potentially different data characteristics, we also include out-of-domain datasets (ECG200 \citep{ECG200} and StandWalkJump \citep{StandWalkJump}). Performance on these datasets, especially with \textbf{limited data for adaptation}, will demonstrate \name's \textbf{utility under data insufficiency}.

\xhdr{Baselines}
We compare \name with 15 SOTA baselines including 11 TSM and 4 TSA models.

The TSM models---Autoformer \citep{Autoformer}, Crossformer \citep{Crossformer}, FEDformer \citep{FEDformer}, Informer \citep{Informer}, iTransformer \citep{iTransformer}, MTST \citep{MTST}, Nonformer \citep{Nonformer}, PatchTST \citep{PatchTST}, Reformer \citep{Reformer}, Transformer \citep{Transformer} and Medformer \citep{Medformer}---are trained independently on each dataset. They represent the conventional approach, used to verify the benefits of \name's repurposing stage in learning transferable domain knowledge.

The TSA models aim to share knowledge across tasks. \textit{TimesFM-TSA} is a key baseline, created by adapting the same TimesFM \citep{TimesFM} backbone with a task-specific CNN head for classification. This allows for a direct evaluation of the additional benefits provided by \name's novel classifier design and the two-stage repurposing/adapting strategy over a more straightforward adaptation of the same foundation model. The additional TSA models, \textit{PatchTST-TSA}, \textit{Reformer-TSA}, and \textit{iTransformer-TSA}, are variants of their TSM counterparts. Their backbones are shared across tasks with task-specific heads, and they are configured to have a similar parameter amount to \name for demonstrating the superiority of using pre-trained foudnation models over training from scratch.

\xhdr{Evaluations}
The effectiveness of our method is primarily demonstrated through its classification performance on the test sets, using metrics such as accuracy, precision, recall, F1 score, AUROC, and AUPRC. Strong performance on these patient-independent test sets will underscore \name's ability to overcome intra-dataset heterogeneity. The generalization ability to unseen tasks, particularly under conditions of potential data insufficiency and inter-dataset heterogeneity, is evaluated through adapting experiments on the small, out-of-domain datasets. These experiments, crucial for assessing the efficacy of the adapting stage where only minimal parameters (CEs and LQs) are learned, are conducted on five random seeds for all models, with results averaged to ensure reliability.

\subsection{Evaluation on Repurposing: Generalize to Unseen Subjects}

\begin{figure}[tb!]
    \centering
    \vskip -0.1in
    \includegraphics[width=\textwidth, trim={0 0.25cm 0 0.25cm}, clip]{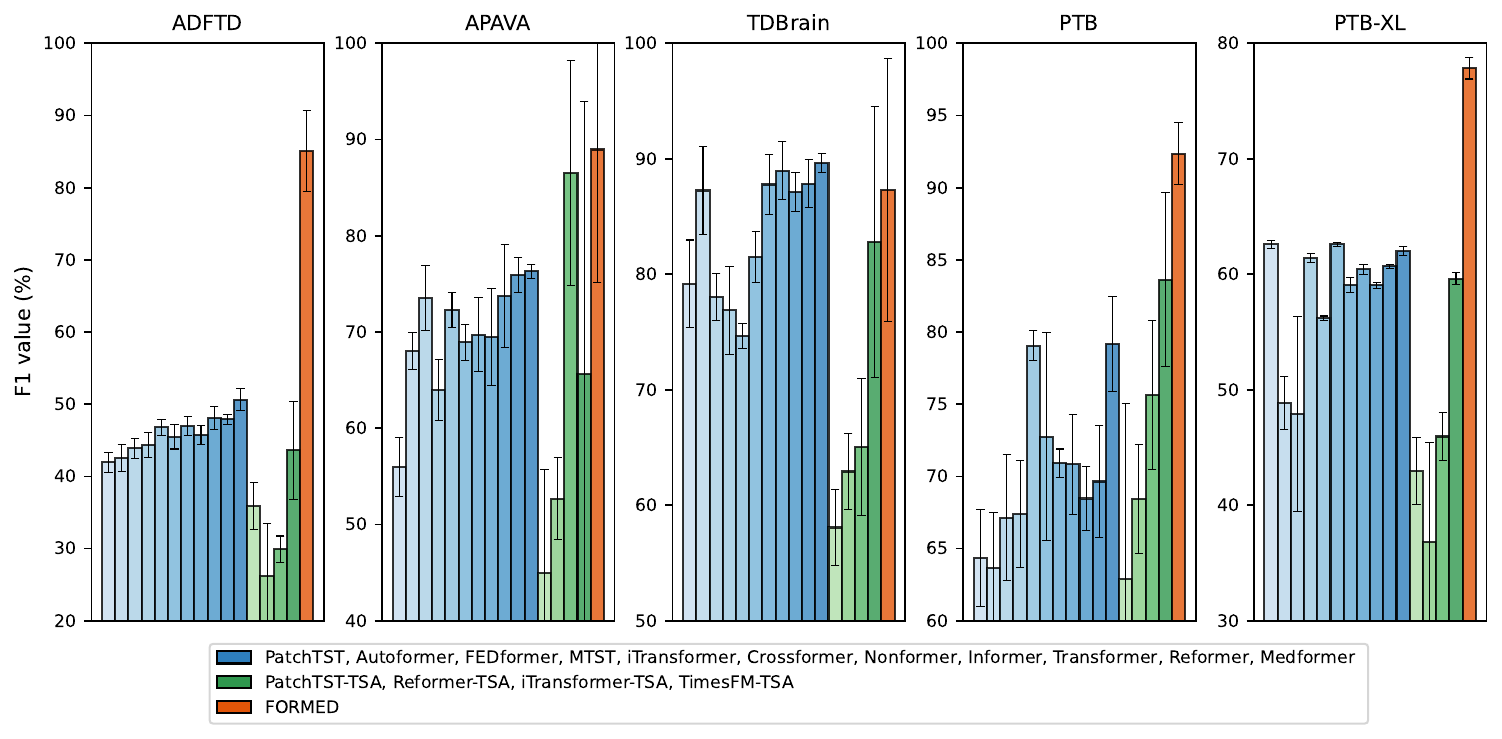}
    \vskip -0.1in
    \caption{In-domain F1 performance on the MedTS cohort datasets. \name achieves SOTA level performance across all datasets in all metrics. Numerical results are shown in \cref{tab:full-result}. Other metrics are included in \cref{sec:experiment-results}.}
    \label{fig:comparison-f1}
    \vskip -0.15in
\end{figure}

\xhdr{Setup}
For repurposing datasets in MedTS cohort, we trained 55 TSM models (11 models for each), and 4 TSA models with 5 task-specific heads each. Our \name model is trained on all 5 datasets, using a fixed $k=16$ for all datasets.

\xhdr{Results}
The results compellingly demonstrate that the proposed repurposing, which allows the SDA to capture shared MedTS domain knowledge while CEs and LQs handle task-specifics, is more effective than both TSM and TSA approaches for complex classification tasks. As shown in \cref{fig:comparison-f1}, \name achieves significant improvements over all baselines. This is particularly pronounced on medium-to-large datasets, such as ADFTD, PTB, and PTB-XL, where performance gains up to 30-40\%. On the relatively small TDBrain dataset, which represents a simpler task where TSMs' performance also saturate, \name performs on par with the strongest TSMs while maintaining a clear advantage over TSA methods. This superiority stems from \name's ability to learn from the collective knowledge of the diverse MedTS cohort, demonstrating that SDA learns generalizable patterns relevant to MedTS classification. Such robust learning across varied datasets translates directly into \textbf{addressing intra-dataset heterogeneity}, \ie, better generalization to unseen subjects.

Interestingly, TSA models generally underperform compared to TSMs, and significantly to \name. This might suggest that their simpler task-specific heads on a shared backbone are less effective at navigating the substantial inter-dataset heterogeneity within the MedTS cohort than dedicated SDA, CE and LQ architecture. TimesFM-TSA, with the same backbone as \name, provides the clearest comparison to this hypothesis.

\subsection{Evaluation on Adapting: Generalize to Unseen Task}

\begin{figure}[tb!]
    \centering
    \includegraphics[width=\textwidth, trim={0 0.05cm 0 0.25cm}, clip]{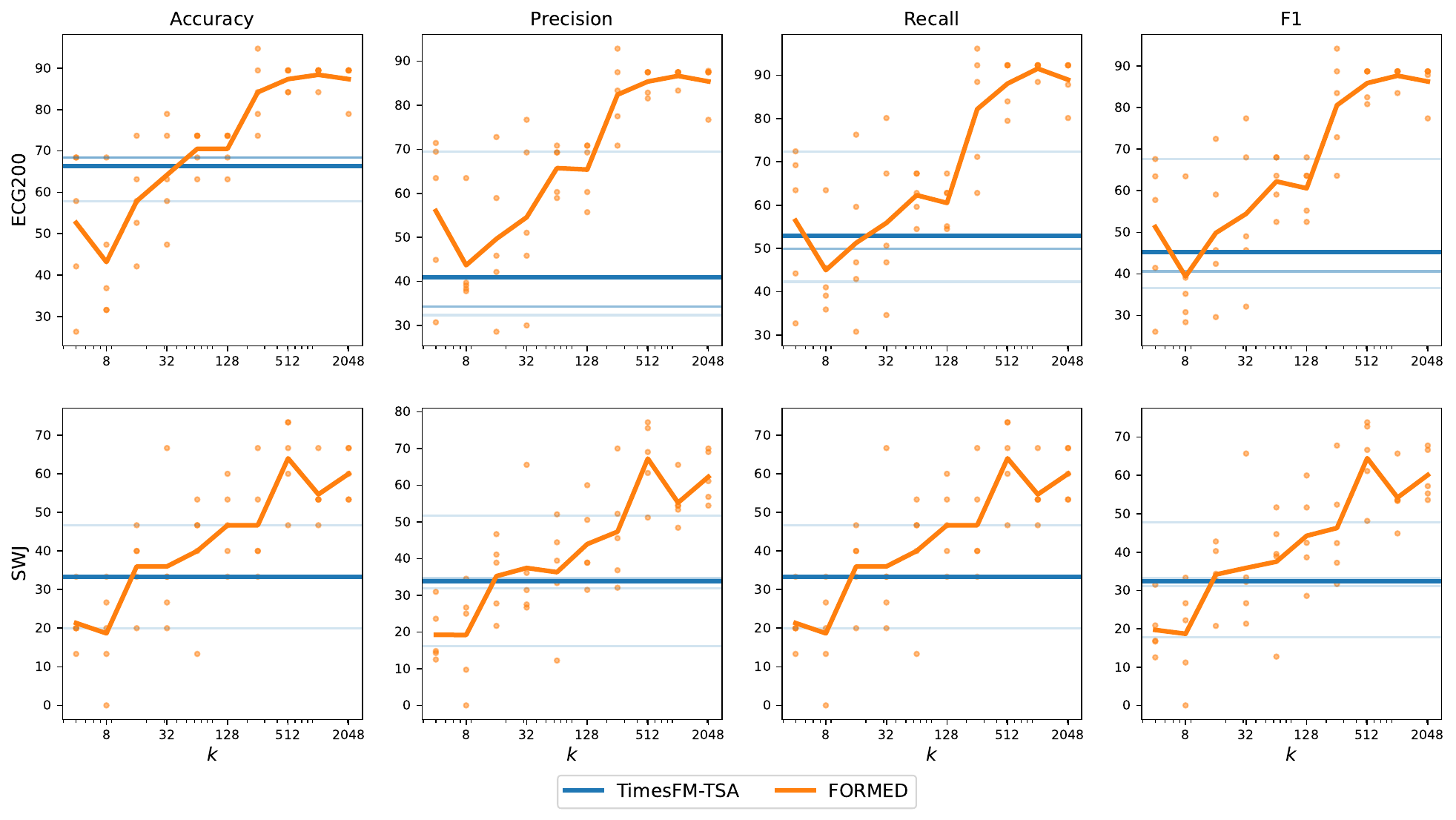}
    \vskip -0.1in
    \caption{Adapt-time scaling on unseen, out-of-domain dataset. \name's performance scales well with $k$ following power law, and outperforms TimesFM-TSA starting from $k=64$ on ECG200, and from $k=16$ on StandWalkJump. Numerical results see \cref{tab:adaptation}.}
    \label{fig:adapting}
    \vskip -0.15in
\end{figure}

\xhdr{Setup}
In this evaluation, we assess \name's ability to generalize to entirely new, unseen tasks, a critical test of its adapting stage and its robustness to \textbf{inter-dataset heterogeneity} and \textbf{data insufficiency}. We use the TimesFM-TSA model as a strong baseline. The \name model is obtained from the repurposing stage; its backbone and SDA parameters are kept frozen. Only newly initialized CEs and LQs are trained for these new tasks, demonstrating \textbf{data-efficient adaptation}. We also explore how performance scales by varying the number of queries per class ($k$) from $4$ to $2048$.

\xhdr{Results}
The TimesFM-TSA baseline, lacking a sophisticated pre-adaptation to the MedTS domain for its classification components, can struggle to generalize from limited data on new tasks, often overfitting to the training data. In contrast, \name, by leveraging the rich MedTS domain knowledge captured in its frozen SDA during the repurposing stage, demonstrates superior adaptation. As seen in \cref{fig:adapting}, \name generally outperforms the TimesFM-TSA baseline on these unseen datasets, even with less adaptable parameters (\name at $k=16$ outperforms TimesFM-TSA with only $\sim$ 1/6 of the parameter). The performance of \name steadily improves with an increase in $k$. This shows that \name can effectively utilize increased capacity within its minimal adaptable components (CEs and LQs) to better handle new tasks without requiring extensive retraining or large datasets. This scalability and efficiency in the adapting stage validates \name's \textbf{strength against data insufficiency and inter-dataset heterogeneity}, making it particularly well-suited for real-world clinical applications where new diagnostic tasks may emerge with \textbf{limited available data}.

%% file: sections/05-conclusion.tex
\section{Discussions and Conclusion}
\label{sec:conclusion}

In this paper, we introduced \textbf{\name}, a novel framework that repurposes general time series foundation models for robust and adaptable MedTS classification. \name's core architectural innovation lies in its attention-based classifier, featuring SDA, CEs, and LQs. This design uniquely equips models to \textbf{handle variable input lengths, diverse channel configurations, and dynamic numbers of output classes}---addressing limitations of conventional TSM and TSA approaches.

Our comprehensive experiments demonstrate \name's strong generalization capabilities, achieving \textbf{state-of-the-art performance for unseen patients within datasets (intra-dataset) and effectively adapting to unseen tasks (inter-dataset).} This highlights two significant findings: first, the feasibility and effectiveness of leveraging powerful pre-trained foundation models as backbones for complex MedTS classification. Second, it validates the superiority of our \name repurposing framework. The framework excels at capturing transferable MedTS domain knowledge within its shared components during an initial repurposing stage, which then \textbf{enables highly efficient and data-scarce adaptation to new tasks by learning only minimal, task-specific parameters.}

Despite these promising results, we acknowledge limitations. Our validation primarily utilized TimesFM as the backbone. While this serves as a strong proof-of-concept for the \name framework's efficacy, its performance and interaction with other diverse time series foundation models warrant future investigation. Additionally, the composition and scale of the MedTS cohort employed during the repurposing stage may also influence the breadth of the captured domain knowledge.

In conclusion, \name presents a significant step towards more \textbf{generalizable}, \textbf{adaptable}, and \textbf{data-efficient} deep learning solutions for the unique challenges posed by medical time series analysis, offering a robust framework for future advancements in the field.

%% file: sections/99-appendix.tex
\section{Comparison of Adaptation Techniques in Foundation Models}
\label{sec:adaptation-comparison}


As discussed in \cref{sec:related-work}, adaptation techniques for foundation models mainly includes \textit{Prompting}, \textit{Fine-tuning}, \textit{Re-programming}, and \textit{Re-purposing}. We have introduced re-programming and re-purposing, and here we provide a brief overview of prompting and fine-tuning, and compare these techniques based on three aspects: \textit{Data Efficiency}, \textit{New Task Type}, and \textit{Generalizability}.

\textit{Prompting \& Fine-tuning}: Both are common adaptation techniques for foundation models, where prompting involves conditioning the model with specific instructions or cues, either handcrafted \citep{Prompt-Engineering,Prompt-Programming} or learned through data \citep{Prompt-Learning}, and fine-tuning involves updating the model's internal parameters on dedicated dataset \citep{Fine-tuning,Delta-Tuning}. While they focus on different aspects of adaptation, they share the commonality of not altering the model's core architecture, therefore the functionality of the model remains unchanged, \eg, model for forecasting remains a forecasting model. Moreover, fine-tuning is often more data-intensive, as it necessitates updating the entire model's parameters, whereas prompting typically only requires learning a limited set of task-specific embeddings or prompts.

In general, these techniques can be categorized based on three aspects: \textit{Data efficiency}, as the scale of dataset used for adaptation, typically measured by the number of parameters updated; \textit{New Task Type}, as the ability to adapt to new tasks that are different from the original task, such as from forecasting to classification; and \textit{Generalizability}, as the ability for the adapted model to be used on unseen datasets and share knowledge across tasks. \cref{tab:related-work} provides a comparison of these techniques based on these aspects.

\input{tables/related-work}

\section{Implementation Details}
\label{sec:implementation}

We take TimesFM \citep{TimesFM} as the backbone for repurposing based on our preliminary comparative analysis of existing time series foundation models.
TimesFM is pre-trained on a largest-scale dataset of diverse time series data for forecasting tasks and is able to capture general time series patterns within dynamic length of historical input. To repurpose it for MedTS classification, we can break down the model's anatomy into three parts, the input patching network, the stacked Transformer, and the output prediction network.

\textbf{Input Patching Network.} Given a univariate time series input $\vx \in \R^T$ and binary mask $\vm \in \{0,1\}^T$ with length $T$, they are first broken up into patches $\mX \in \R^{L \times P}$ and $\mM \in \{0,1\}^{L \times P}$ in a non-overlapping fashion, where $P$ is the patch size and $L = \ceil{\frac{T}{P}}$ is the number of tokens. Each patch $\mX_{i,:}$ is the concatenation of $P$ consecutive elements of the input sequence $\vx$ in a non-overlapping fashion and so is the $\mM_{i,:}$. The $\mX_{i,:}$ and $\mM_{i,:}$ denote the $i$-th row of $\mX$ and $\mM$, respectively. The sequence of patches $\mX$ and $\mM$ are then projected to a sequence of tokens $\mZ \in \R^{L \times D}$ in the model dimension $D$ using an input residual block:
\begin{equation}
    \begin{aligned}
        \mZ_{i,:} = & \texttt{InputResidualBlock}(\mX_{i,:}; \mM_{i,:})
    \end{aligned}
\end{equation}

\textbf{Stacked Transformer.} Before passing into the stacked Transformer, the positional encoding will be added to the tokens to form the input sequence $\tilde\mZ \in \R^{L \times D}$. The stacked Transformer is then applied to the input sequence $\tilde\mZ$ to capture the temporal dependencies and extract features using casual self-attention, outputting feature rich tokens $\mH\in\R^{L\times D}$:
\begin{equation}
    \begin{aligned}
        \tilde\mZ_{i,:} = & \mZ_{i,:} \oplus \texttt{PositionalEncoding}(i)                                                                       \\
        \mH_{i,:} =       & \texttt{StackedTransformer}(\tilde\mZ_{1,:},\tilde\mZ_{2,:},...,\tilde\mZ_{i,:};\dot\evm_1,\dot\evm_2,...,\dot\evm_i)
    \end{aligned}
\end{equation}
where $\dot\evm_i = \min\{\mM_{i,:}\}$ is the mask for the $i$-th patch for masking out completely empty ones.

\textbf{Output Prediction Network.} The output prediction network is a residual block layer that maps the last output $\mH_{L,:}$ from the Transformer back to the original input spaces $\hat\vx \in \R^N$, forming the prediction of the next $N$ time steps:
\begin{equation}
    \begin{aligned}
        \hat\vx = \texttt{OutputResidualBlock}(\mH_{L,:})
    \end{aligned}
\end{equation}

In summary, the duty of prediction lies solely on the last output prediction network, while the input patching network plus the stacked Transformer can be viewed as a feature extractor that maps the input time series $\vx$ to a sequence of feature tokens $\mH$ (\cref{fig:architecture}). This can be easily extended to process multivariate MedTS by processing each channel of input individually and stack the extracted features as $\tH \in \R^{C \times L \times D}$ for data of $C$ channels. This will serve as the backbone feature extractor for the downstream classification model.

\section{Experiment Setup}
\label{sec:experiment-setup}

\subsection{Experimental Setup}
The experiments are ran on several hosts. The following list summarizes the hardware and software configurations used in our experiments:
\input{tables/environment}

\subsection{Code Availability}
The code will be available upon acceptance.

\subsection{Dataset Availability \& Preprocessing}
The preprocessed datasets in the MedTS cohort are obtained from the authors of \citet{Medformer}. The datasets used for adapting are loaded through \texttt{sktime} API and no additional preprocessing is performed.

\subsection{Data Splitting}
For datasets in the MedTS cohort, the datasets are splitted into training, validation, and test sets in the ratio of 6:2:2 at patient level according to \citet{Medformer,PatientIndependent}. The adapting datasets are not clearly defined in terms of patient, so we perform best effort to split the datasets into training, validation, and test sets in the ratio of 8:1:1 at recording level. The splitting is seeded and stratified to ensure that the training, validation, and test sets have similar distributions of the target classes. The training set is used for training the model, the validation set is used for early stopping, and the test set is used for evaluating the model performance.

\subsection{Model Training}
\begin{figure*}[!htb]
    \centering
    \begin{minipage}{0.5\textwidth}
        \includegraphics[width=\textwidth, trim={0.25cm 0 0 0.25cm}, clip]{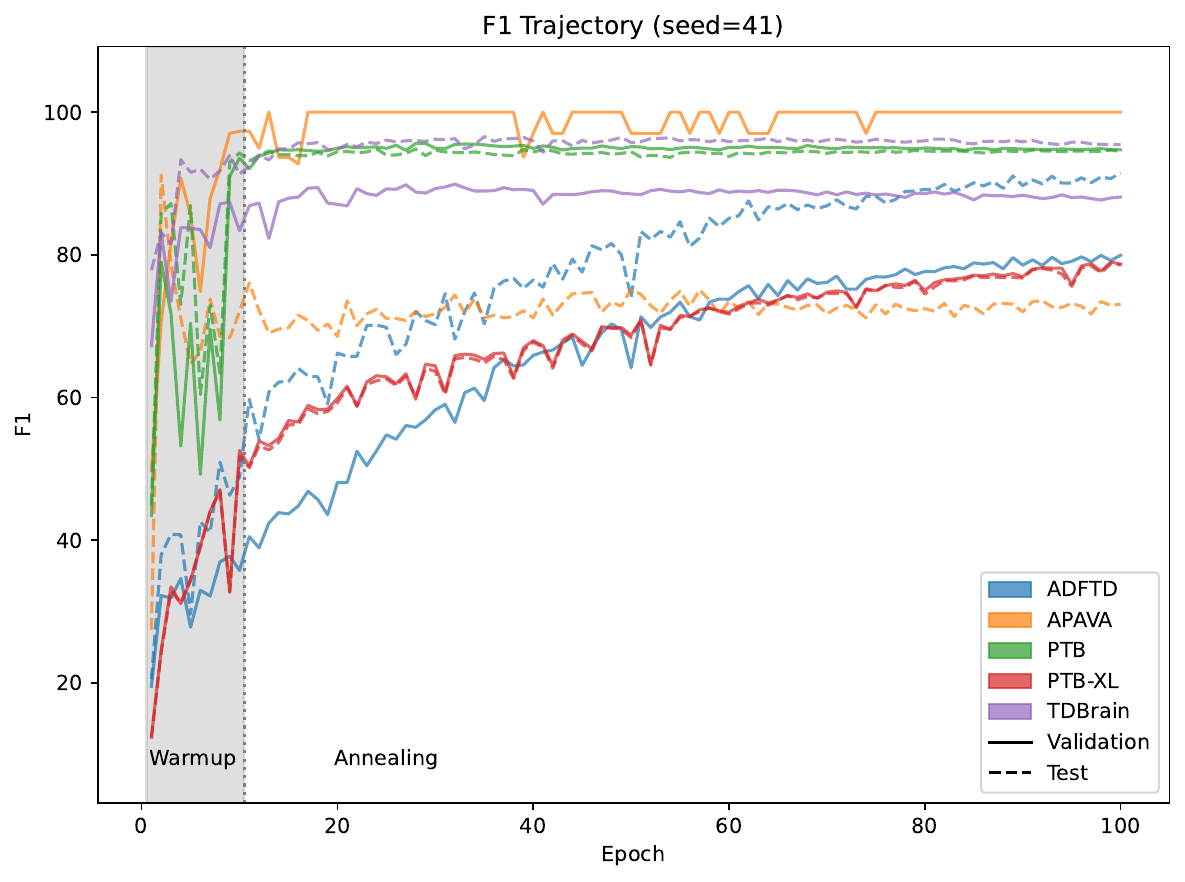}
    \end{minipage}%
    \begin{minipage}{0.5\textwidth}
        \includegraphics[width=\textwidth, trim={0.25cm 0 0 0.25cm}, clip]{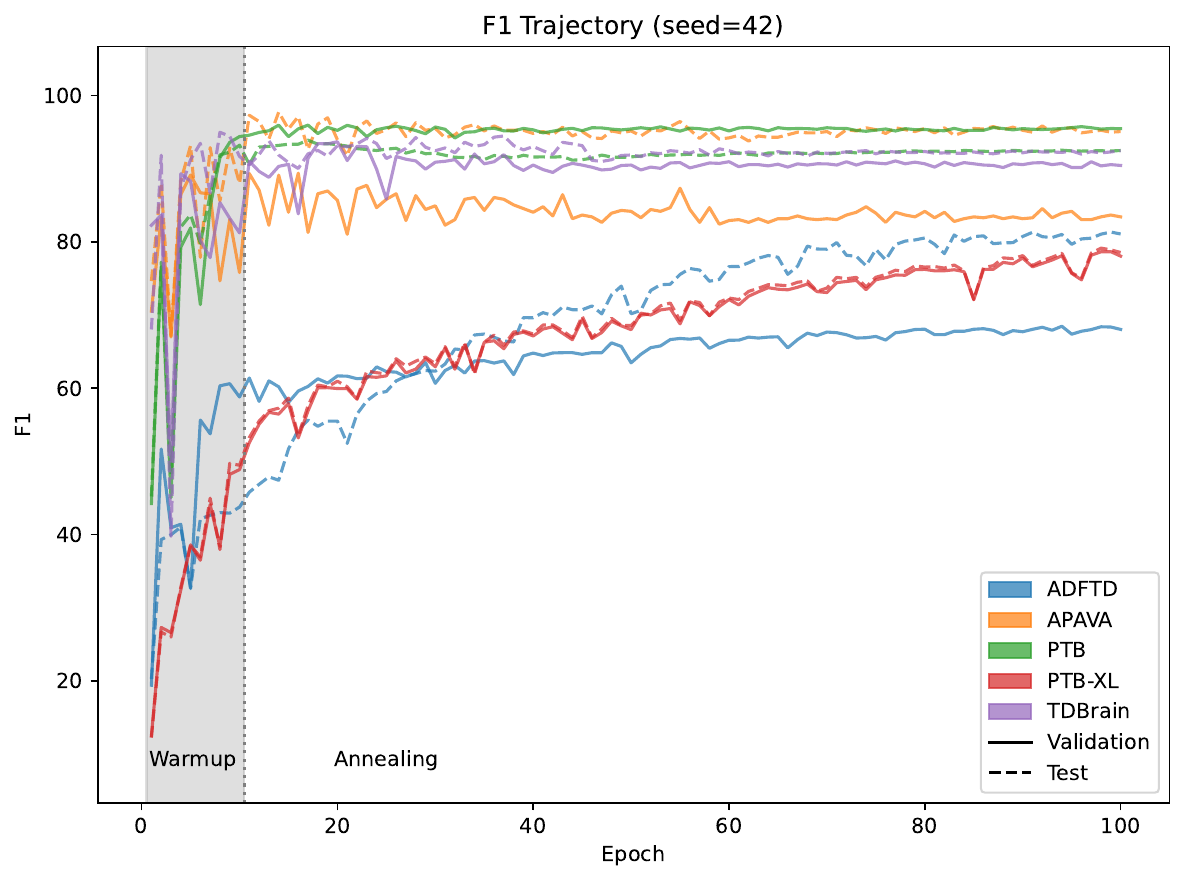}
    \end{minipage}
    \caption{Typical training trajectories of \name during repurposing from two different seeds.}
    \label{fig:metric-trajectory}
\end{figure*}

The frozen pre-trained backbone TimesFM model is the v2 version from the official GitHub repository, with 50 layers and model dimension of 1280, and patch size of 32. The classifier in \name is initialized with default initialization method, and the embeddings are initialized to normal distribution with $\mu=0$ and $\sigma^2=0.1$. The model is trained with \texttt{AdamW} optimizer with weight decay of \num{1e-3} and a custom log-normal learning rate schedule. For each epoch $t$, the learning rate is calculated as:
\begin{equation}
    \begin{aligned}
        \mathcal{LN}(t;\mu,\sigma^2,T) = &
        \begin{cases}
            0                                                                                           & t = 0            \\
            \frac{T}{t} \cdot \exp\left(-\frac{\left(\ln{t} - \ln{T} - \mu\right)^2}{2 \sigma^2}\right) & \text{otherwise}
        \end{cases}                                                                                 \\
        \texttt{lr}(t;\mu,\sigma^2,T) =  & \texttt{lr}_{\min} + \frac{\mathcal{LN}(t;\mu,\sigma^2,T)}{\max_t{\mathcal{LN}(t;\mu,\sigma^2,T)}}\cdot\left(\texttt{lr}_{\max} - \texttt{lr}_{\min}\right)
    \end{aligned}
\end{equation}
where we use $\mu=1$, $\sigma^2=1$, $T=10$, $\texttt{lr}_{\min}=\num{1e-5}$ and $\texttt{lr}_{\max}=\num{1e-3}$ in our experiments. This creates a quick warm-up phase followed by a gradual decay of the learning rate for annealing. The \cref{fig:metric-trajectory} shows the typical training trajectories of \name during repurposing.

During each epoch of training, the model is trained on 100 batches of data from each dataset sampled at random order. The batch size is tailored for each dataset according to the available samples in each dataset, making the 100 batches of data approximately equal to one effective iteration of the whole training set. Additionally, gradient clipping of $1.0$ is applied to prevent exploding gradients and overshooting in earlier epochs. The model is trained for 100 epochs with early stopping on the validation set based on the average F1-score with patience of 10 epochs. The same training procedure is applied to all TSA and our method.

\subsection{Hyperparameter Tuning}
The initial hyperparameter $k$ for the classifier in \name was not tuned due to the high cost associated with repurposing extra-large models like TimesFM, but rather set to an arbitrary value of $k=16$ for all datasets. However, we do perform a grid search for the hyperparameter $k$ during the adapting process, as shown in \cref{fig:adapting}. We initiate the search with powers of $4$, until the performance shows no significant improvement. Then we perform a finer search with powers of $2$. We find that $k=256\sim1024$ is a good range for adapting to small datasets. This search is very efficient as the computation cost of increasing $k$ is often negligible compared to the computation cost of large model backbones, despite being linear in $k$ theoretically.

\section{Datasets}
\label{sec:dataset-detail}

Here we provide the details of the datasets \cref{tab:dataset} used as the MedTS cohort for repurposing in \cref{sec:experiment}. The datasets are publicly available, and we follow the pre-processing and splitting procedures as in \citet{Medformer}.

\input{tables/dataset}

\section{Experimental Results}
\label{sec:experiment-results}

\begin{figure}[htb!]
    \centering
    \vskip -0.1in
    \includegraphics[width=\textwidth, trim={0 0.25cm 0 0.25cm}, clip]{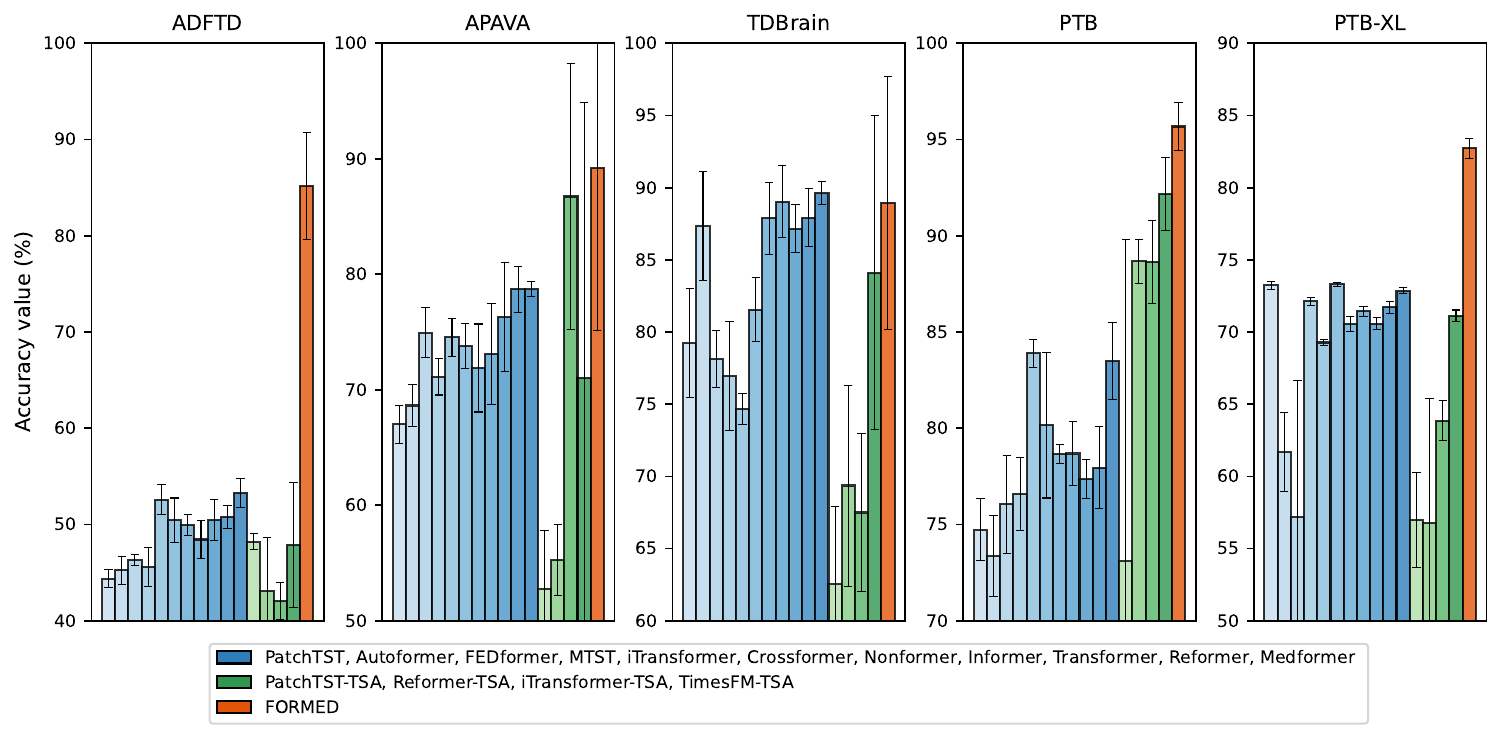}
    \vskip -0.1in
    \caption{In-domain accuracy performance on the MedTS cohort datasets.}
    \label{fig:comparison-accuracy}
    \vskip -0.15in
\end{figure}

\begin{figure}[htb!]
    \centering
    \vskip -0.1in
    \includegraphics[width=\textwidth, trim={0 0.25cm 0 0.25cm}, clip]{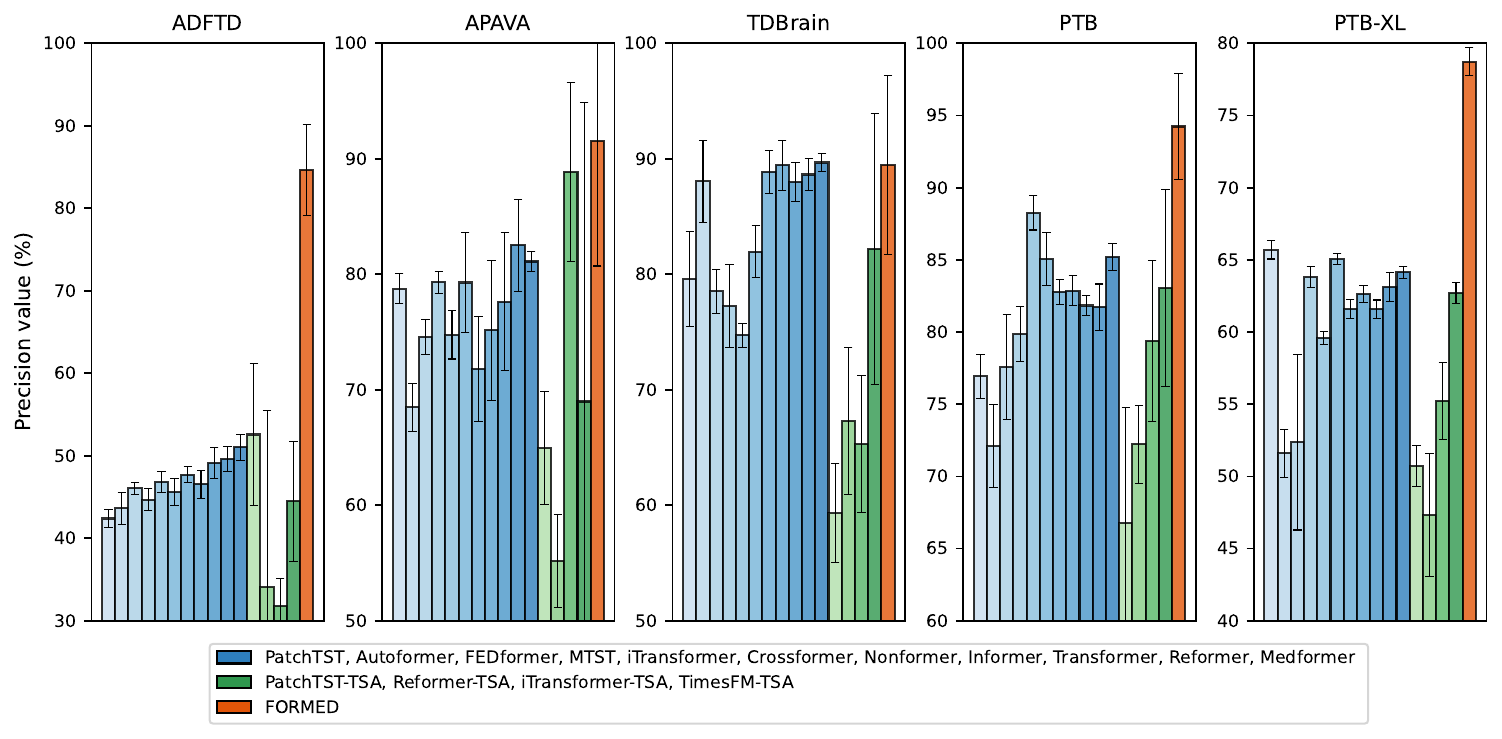}
    \vskip -0.1in
    \caption{In-domain precision performance on the MedTS cohort datasets.}
    \label{fig:comparison-precision}
    \vskip -0.15in
\end{figure}

\begin{figure}[htb!]
    \centering
    \vskip -0.1in
    \includegraphics[width=\textwidth, trim={0 0.25cm 0 0.25cm}, clip]{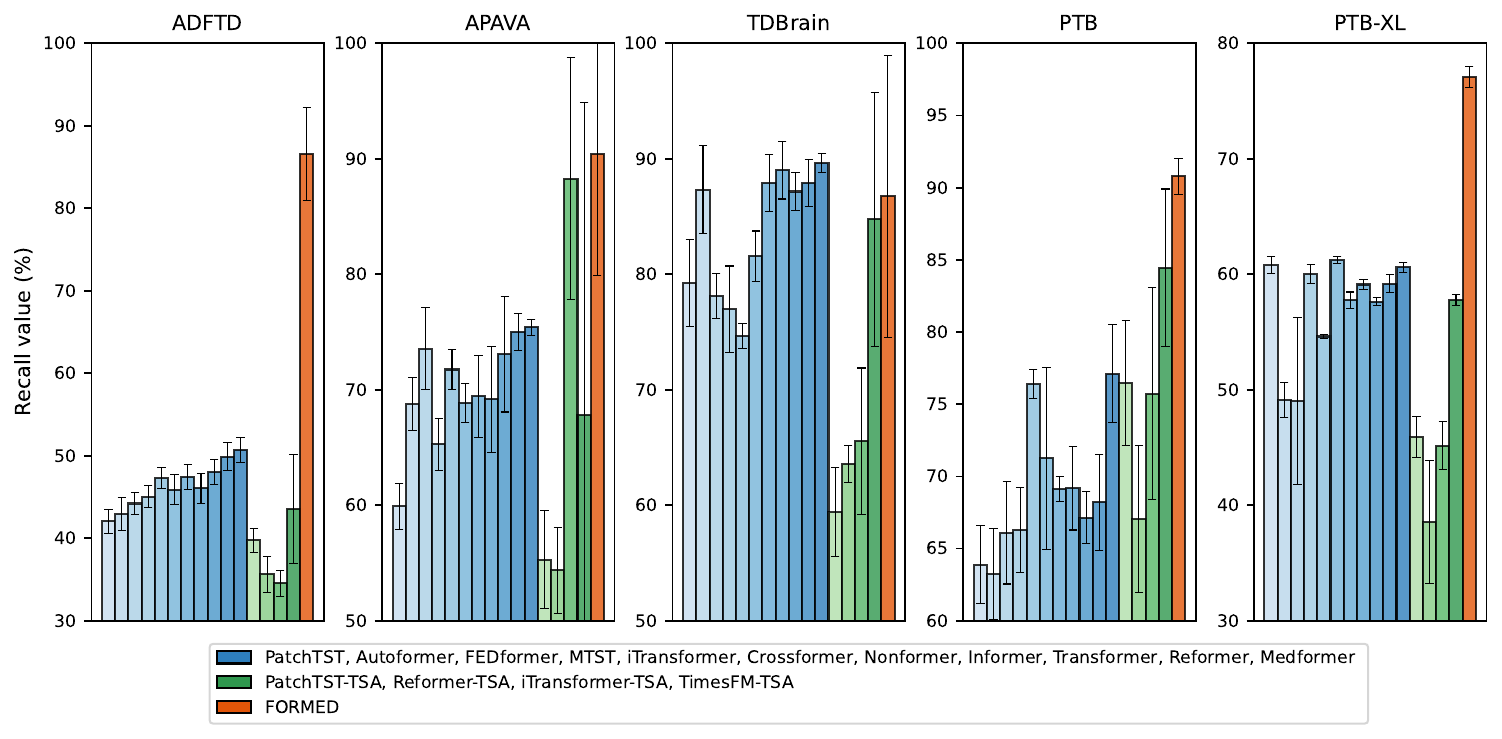}
    \vskip -0.1in
    \caption{In-domain recall performance on the MedTS cohort datasets.}
    \label{fig:comparison-recall}
    \vskip -0.15in
\end{figure}

\begin{figure}[htb!]
    \centering
    \vskip -0.1in
    \includegraphics[width=\textwidth, trim={0 0.25cm 0 0.25cm}, clip]{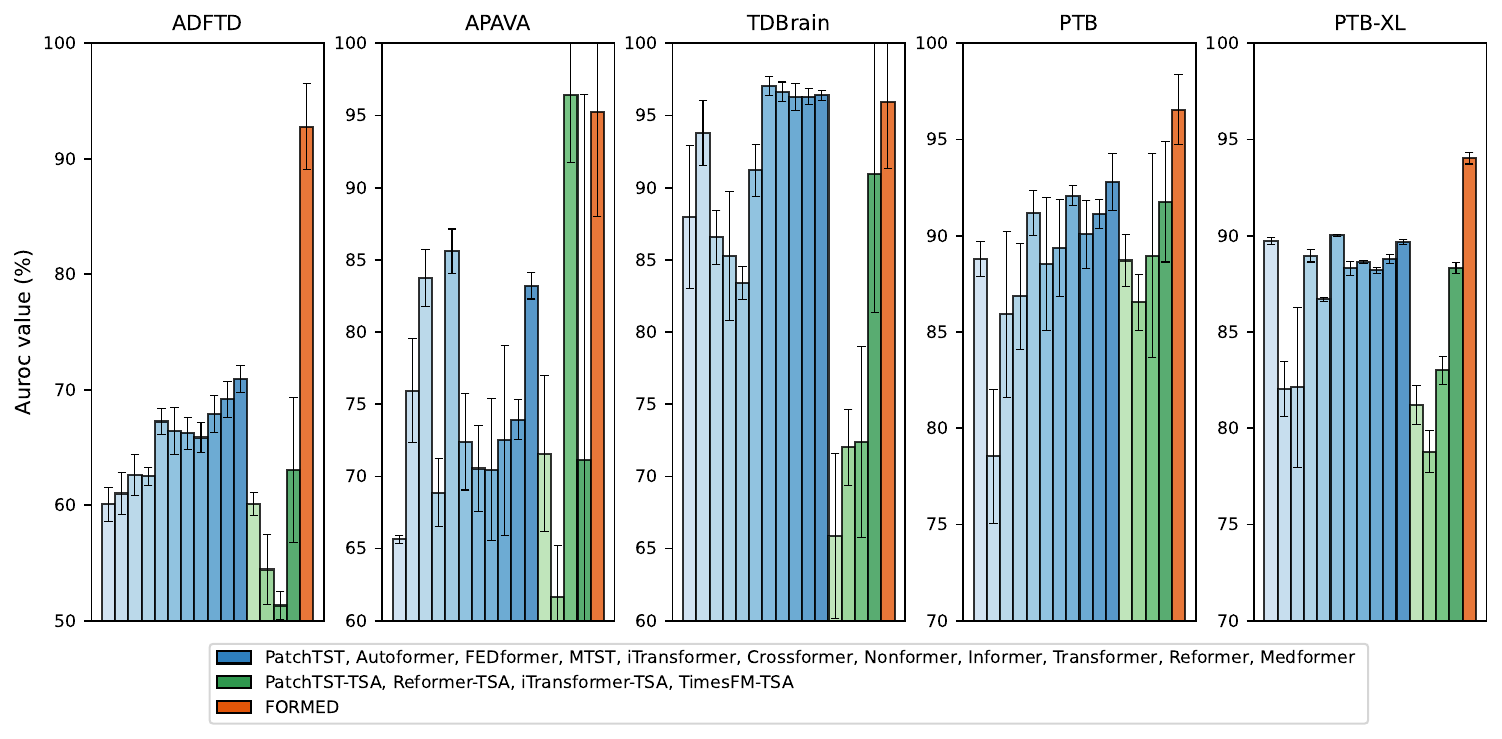}
    \vskip -0.1in
    \caption{In-domain AUROC performance on the MedTS cohort datasets.}
    \label{fig:comparison-auroc}
    \vskip -0.15in
\end{figure}

\begin{figure}[htb!]
    \centering
    \vskip -0.1in
    \includegraphics[width=\textwidth, trim={0 0.25cm 0 0.25cm}, clip]{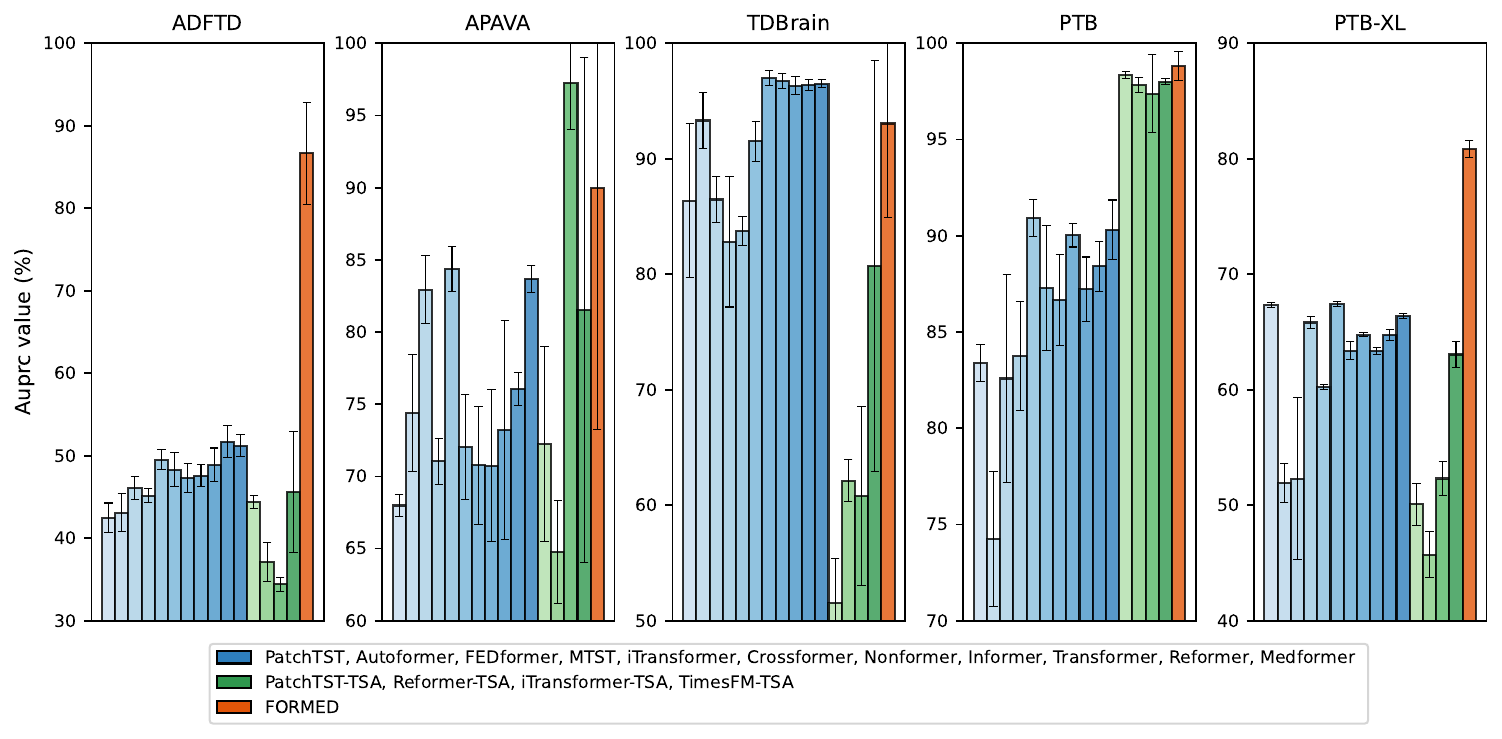}
    \vskip -0.1in
    \caption{In-domain AUPRC performance on the MedTS cohort datasets.}
    \label{fig:comparison-auprc}
    \vskip -0.15in
\end{figure}

\input{tables/full-comparison}

\input{tables/adapting}

\clearpage

%% file: tables/related-work.tex
\begin{table}[h!]
    \centering
    \vskip -0.15in
    \def\arraystretch{1.0}
    \caption{Comparison of adaptation techniques of time series foundation models.
    }
    \label{tab:related-work}
    \begin{tabular}{cccc}
        \toprule
        \textbf{Adaptation} & \textbf{Data Efficiency} & \textbf{New Task Type} & \textbf{Generalizability} \\
        \midrule
        Prompting           & \checkmark               &                        & \checkmark\footnotemark   \\
        Fine-tuning         &                          &                        & \checkmark                \\
        Re-programming      &                          & \checkmark             &                           \\
        \rowcolor{yellow!50}
        Re-purposing        & \checkmark               & \checkmark             & \checkmark                \\
        \bottomrule
    \end{tabular}
    \vskip -0.1in
\end{table}

\addtocounter{footnote}{-1}
\stepcounter{footnote}\footnotetext{Although the model structure is fixed and still applicable to other datasets and tasks, the engineered or learned prompts can be task-specific.}

%% file: tables/environment.tex
\begin{table}[!htb]
    \centering
    \def\arraystretch{1.0}
    \caption{\textbf{Environment setup}.}
    \label{tab:environment}
    \resizebox{0.9\textwidth}{!}{%
        \begin{tabular}{cccccccc}
            \toprule
            \textbf{Host No.} & \textbf{CPU}                      & \textbf{Memory (GB)} & \textbf{GPU}                \\
            \midrule
            1                 & Intel Core i9-10900X              & \num{32}             & NVIDIA RTX A5000 $\times$ 1 \\
            2                 & AMD Ryzen Threadripper PRO 3995WX & \num{512}            & NVIDIA RTX A5000 $\times$ 4 \\
            3                 & AMD EPYC 7713                     & \num{1024}           & NVIDIA RTX A5000 $\times$ 8 \\
            4                 & AMD EPYC 7513                     & \num{256}            & NVIDIA RTX A6000 $\times$ 8 \\
            \bottomrule
        \end{tabular}
    }
    \resizebox{0.35\textwidth}{!}{%
        \begin{tabular}{cccccccc}
            \toprule
            \textbf{Software/Package} & \textbf{Version} \\
            \midrule
            Python3                   & 3.13.3           \\
            PyTorch                   & 2.7.0            \\
            CUDA                      & 12.4 -- 12.6     \\
            \bottomrule
        \end{tabular}
    }
    \vskip -0.1in
\end{table}

%% file: tables/dataset.tex
\begin{table}[!ht]
    \centering
    \def\arraystretch{1.0}
    \caption{\textbf{MedTS Cohort Datasets}.    }
    \label{tab:dataset}
    \resizebox{0.9\textwidth}{!}{%
        \begin{tabular}{cccccccc}
            \toprule
            \textbf{Dataset}              & \textbf{Type} & \textbf{\# Subject} & \textbf{\# Sample} & \makecell{\textbf{Sampling}                                  \\\textbf{Rate}} & \makecell{\textbf{Sampling}\\\textbf{Length}} & \textbf{\# Channel} & \textbf{\# Classes} \\
            \midrule
            ADFTD \citep{ADFTD-1,ADFTD-2} & EEG           & \num{88}            & \num{69762}        & \qty{256}{\hertz}           & \num{256} & \num{19} & \num{3} \\
            APAVA \citep{APAVA}           & EEG           & \num{23}            & \num{5967}         & \qty{256}{\hertz}           & \num{256} & \num{16} & \num{2} \\
            TDBrain \citep{TDBrain}       & EEG           & \num{72}            & \num{6240}         & \qty{256}{\hertz}           & \num{256} & \num{33} & \num{2} \\
            PTB \citep{PTB}               & ECG           & \num{198}           & \num{64356}        & \qty{250}{\hertz}           & \num{300} & \num{15} & \num{2} \\
            PTB-XL \citep{PTB-XL}         & ECG           & \num{17596}         & \num{191400}       & \qty{250}{\hertz}           & \num{250} & \num{12} & \num{5} \\
            \bottomrule
        \end{tabular}
    }
    \vskip -0.1in
\end{table}

%% file: tables/full-comparison.tex
\begin{table*}[!ht]
  \centering
  \def\arraystretch{1.0}
  \caption{\textbf{Results on MedTS Cohort for disease classification.}
    Best results from non-TSM models are \textbf{bolded} and best results of all models are \underline{underlined}. TimesFM-TSA shows great improvement over other TSA models and achieves competitive performance with the SOTA TSM models.
    While our model, \name, consistently outperforms the all other model on all datasets.
  }
  \label{tab:full-result}
  \resizebox{0.85\textwidth}{!}{%
    \begin{tabular}{cclllllll}
      \toprule
      \textbf{Datasets} & \textbf{Adaptation}   & \textbf{Models}           & \textbf{Accuracy}                     & \textbf{Precision}                    & \textbf{Recall}                       & \textbf{F1 score}                     & \textbf{AUROC}                       & \textbf{AUPRC}                       \\
      \midrule

      \multirow{16}{*}{\makecell{\textbf{ADFTD}                                                                                                                                                                                                                                                                           \\ (3-Classes)}}
                        & \multirow{11}{*}{TSM} & \textbf{Autoformer}       & 45.25\std{1.48}                       & 43.67\std{1.94}                       & 42.96\std{2.03}                       & 42.59\std{1.85}                       & 61.02\std{1.82}                      & 43.10\std{2.30}                      \\
                        &                       & \textbf{Crossformer}      & 50.45\std{2.31}                       & 45.57\std{1.63}                       & 45.88\std{1.82}                       & 45.50\std{1.70}                       & 66.45\std{2.03}                      & 48.33\std{2.05}                      \\
                        &                       & \textbf{FEDformer}        & 46.30\std{0.59}                       & 46.05\std{0.76}                       & 44.22\std{1.38}                       & 43.91\std{1.37}                       & 62.62\std{1.75}                      & 46.11\std{1.44}                      \\
                        &                       & \textbf{Informer}         & 48.45\std{1.96}                       & 46.54\std{1.68}                       & 46.06\std{1.84}                       & 45.74\std{1.38}                       & 65.87\std{1.27}                      & 47.60\std{1.30}                      \\
                        &                       & \textbf{iTransformer}     & 52.60\std{1.59}                       & 46.79\std{1.27}                       & 47.28\std{1.29}                       & 46.79\std{1.13}                       & 67.26\std{1.16}                      & 49.53\std{1.21}                      \\
                        &                       & \textbf{MTST}             & 45.60\std{2.03}                       & 44.70\std{1.33}                       & 45.05\std{1.30}                       & 44.31\std{1.74}                       & 62.50\std{0.81}                      & 45.16\std{0.85}                      \\
                        &                       & \textbf{Nonformer}        & 49.95\std{1.05}                       & 47.71\std{0.97}                       & 47.46\std{1.50}                       & 46.96\std{1.35}                       & 66.23\std{1.37}                      & 47.33\std{1.78}                      \\
                        &                       & \textbf{PatchTST}         & 44.37\std{0.95}                       & 42.40\std{1.13}                       & 42.06\std{1.48}                       & 41.97\std{1.37}                       & 60.08\std{1.50}                      & 42.49\std{1.79}                      \\
                        &                       & \textbf{Reformer}         & 50.78\std{1.17}                       & 49.64\std{1.49}                       & 49.89\std{1.67}                       & 47.94\std{0.69}                       & 69.17\std{1.58}                      & 51.73\std{1.94}                      \\
                        &                       & \textbf{Transformer}      & 50.47\std{2.14}                       & 49.13\std{1.83}                       & 48.01\std{1.53}                       & 48.09\std{1.59}                       & 67.93\std{1.59}                      & 48.93\std{2.02}                      \\
                        &                       & \textbf{Medformer}        & 53.27\std{1.54}                       & 51.02\std{1.57}                       & 50.71\std{1.55}                       & 50.65\std{1.51}                       & 70.93\std{1.19}                      & 51.21\std{1.32}                      \\
      \cmidrule{2-9}
                        & \multirow{4}{*}{TSA}  & \textbf{iTransformer-TSA} & 42.08\std{1.95}                       & 31.85\std{3.29}                       & 34.53\std{1.58}                       & 29.89\std{1.86}                       & 51.33\std{1.19}                      & 34.42\std{0.84}                      \\
                        &                       & \textbf{PatchTST-TSA}     & 48.23\std{0.80}                       & 52.58\std{8.56}                       & 39.75\std{1.42}                       & 35.87\std{3.27}                       & 60.14\std{1.01}                      & 44.42\std{0.81}                      \\
                        &                       & \textbf{Reformer-TSA}     & 43.09\std{5.61}                       & 34.11\std{21.43}                      & 35.62\std{2.22}                       & 26.21\std{7.22}                       & 54.44\std{3.06}                      & 37.14\std{2.33}                      \\
                        &                       & \textbf{TimesFM-TSA}      & 47.89\std{6.47}                       & 44.50\std{7.29}                       & 43.60\std{6.60}                       & 43.61\std{6.82}                       & 63.07\std{6.27}                      & 45.59\std{7.30}                      \\
      \cmidrule{2-9}
                        & GA                    & \textbf{\name (Ours)}     & \underline{\textbf{85.19\std{5.56}}}  & \underline{\textbf{84.65\std{5.52}}}  & \underline{\textbf{86.61\std{5.63}}}  & \underline{\textbf{85.12\std{5.64}}}  & \underline{\textbf{92.77\std{3.71}}} & \underline{\textbf{86.65\std{6.16}}} \\

      \midrule
      \multirow{16}{*}{\makecell{\textbf{APAVA}                                                                                                                                                                                                                                                                           \\ (2-Classes)}} 
                        & \multirow{11}{*}{TSM} & \textbf{Autoformer}       & 68.64\std{1.82}                       & 68.48\std{2.10}                       & 68.77\std{2.27}                       & 68.06\std{1.94}                       & 75.94\std{3.61}                      & 74.38\std{4.05}                      \\
                        &                       & \textbf{Crossformer}      & 73.77\std{1.95}                       & 79.29\std{4.36}                       & 68.86\std{1.70}                       & 68.93\std{1.85}                       & 72.39\std{3.33}                      & 72.05\std{3.65}                      \\
                        &                       & \textbf{FEDformer}        & 74.94\std{2.15}                       & 74.59\std{1.50}                       & 73.56\std{3.55}                       & 73.51\std{3.39}                       & 83.72\std{1.97}                      & 82.94\std{2.37}                      \\
                        &                       & \textbf{Informer}         & 73.11\std{4.40}                       & 75.17\std{6.06}                       & 69.17\std{4.56}                       & 69.47\std{5.06}                       & 70.46\std{4.91}                      & 70.75\std{5.27}                      \\
                        &                       & \textbf{iTransformer}     & 74.55\std{1.66}                       & 74.77\std{2.10}                       & 71.76\std{1.72}                       & 72.30\std{1.79}                       & 85.59\std{1.55}                      & 84.39\std{1.57}                      \\
                        &                       & \textbf{MTST}             & 71.14\std{1.59}                       & 79.30\std{0.97}                       & 65.27\std{2.28}                       & 64.01\std{3.16}                       & 68.87\std{2.34}                      & 71.06\std{1.60}                      \\
                        &                       & \textbf{Nonformer}        & 71.89\std{3.81}                       & 71.80\std{4.58}                       & 69.44\std{3.56}                       & 69.74\std{3.84}                       & 70.55\std{2.96}                      & 70.78\std{4.08}                      \\
                        &                       & \textbf{PatchTST}         & 67.03\std{1.65}                       & 78.76\std{1.28}                       & 59.91\std{2.02}                       & 55.97\std{3.10}                       & 65.65\std{0.28}                      & 67.99\std{0.76}                      \\
                        &                       & \textbf{Reformer}         & 78.70\std{2.00}                       & 82.50\std{3.95}                       & 75.00\std{1.61}                       & 75.93\std{1.82}                       & 73.94\std{1.40}                      & 76.04\std{1.14}                      \\
                        &                       & \textbf{Transformer}      & 76.30\std{4.72}                       & 77.64\std{5.95}                       & 73.09\std{5.01}                       & 73.75\std{5.38}                       & 72.50\std{6.60}                      & 73.23\std{7.60}                      \\
                        &                       & \textbf{Medformer}        & 78.74\std{0.64}                       & 81.11\std{0.84}                       & 75.40\std{0.66}                       & 76.31\std{0.71}                       & 83.20\std{0.91}                      & 83.66\std{0.92}                      \\
      \cmidrule{2-9}
                        & \multirow{4}{*}{TSA}  & \textbf{iTransformer-TSA} & 86.74\std{11.55}                      & 88.87\std{7.74}                       & 88.28\std{10.49}                      & 86.50\std{11.69}                      & \underline{\textbf{96.39\std{4.66}}} & \underline{\textbf{97.28\std{3.26}}} \\
                        &                       & \textbf{PatchTST-TSA}     & 52.77\std{5.09}                       & 64.96\std{4.92}                       & 55.29\std{4.25}                       & 45.00\std{10.68}                      & 71.56\std{5.40}                      & 72.26\std{6.76}                      \\
                        &                       & \textbf{Reformer-TSA}     & 55.27\std{3.06}                       & 55.20\std{4.01}                       & 54.37\std{3.74}                       & 52.69\std{4.24}                       & 61.64\std{3.60}                      & 64.75\std{3.56}                      \\
                        &                       & \textbf{TimesFM-TSA}      & 70.99\std{23.87}                      & 68.99\std{25.89}                      & 67.83\std{27.01}                      & 65.68\std{28.26}                      & 71.14\std{25.31}                     & 81.52\std{17.49}                     \\
      \cmidrule{2-9}
                        & GA                    & \textbf{\name (Ours)}     & \underline{\textbf{89.20\std{14.07}}} & \underline{\textbf{91.51\std{10.79}}} & \underline{\textbf{90.40\std{10.51}}} & \underline{\textbf{88.97\std{13.85}}} & 95.27\std{7.27}                      & 89.96\std{16.69}                     \\

      \midrule
      \multirow{16}{*}{\makecell{\textbf{TDBrain}                                                                                                                                                                                                                                                                         \\ (2-Classes)}} 
                        & \multirow{11}{*}{TSM} & \textbf{Autoformer}       & 87.33\std{3.79}                       & 88.06\std{3.56}                       & 87.33\std{3.79}                       & 87.26\std{3.84}                       & 93.81\std{2.26}                      & 93.32\std{2.42}                      \\
                        &                       & \textbf{Crossformer}      & 81.56\std{2.19}                       & 81.97\std{2.25}                       & 81.56\std{2.19}                       & 81.50\std{2.20}                       & 91.20\std{1.78}                      & 91.51\std{1.71}                      \\
                        &                       & \textbf{FEDformer}        & 78.13\std{1.98}                       & 78.52\std{1.91}                       & 78.13\std{1.98}                       & 78.04\std{2.01}                       & 86.56\std{1.86}                      & 86.48\std{1.99}                      \\
                        &                       & \textbf{Informer}         & 89.02\std{2.50}                       & 89.43\std{2.14}                       & 89.02\std{2.50}                       & 88.98\std{2.54}                       & 96.64\std{0.68}                      & 96.75\std{0.63}                      \\
                        &                       & \textbf{iTransformer}     & 74.67\std{1.06}                       & 74.71\std{1.06}                       & 74.67\std{1.06}                       & 74.65\std{1.06}                       & 83.37\std{1.14}                      & 83.73\std{1.27}                      \\
                        &                       & \textbf{MTST}             & 76.96\std{3.76}                       & 77.24\std{3.59}                       & 76.96\std{3.76}                       & 76.88\std{3.83}                       & 85.27\std{4.46}                      & 82.81\std{5.64}                      \\
                        &                       & \textbf{Nonformer}        & 87.88\std{2.48}                       & 88.86\std{1.84}                       & 87.88\std{2.48}                       & 87.78\std{2.56}                       & \underline{97.05\std{0.68}}          & \underline{96.99\std{0.68}}          \\
                        &                       & \textbf{PatchTST}         & 79.25\std{3.79}                       & 79.60\std{4.09}                       & 79.25\std{3.79}                       & 79.20\std{3.77}                       & 87.95\std{4.96}                      & 86.36\std{6.67}                      \\
                        &                       & \textbf{Reformer}         & 87.92\std{2.01}                       & 88.64\std{1.40}                       & 87.92\std{2.01}                       & 87.85\std{2.08}                       & 96.30\std{0.54}                      & 96.40\std{0.45}                      \\
                        &                       & \textbf{Transformer}      & 87.17\std{1.67}                       & 87.99\std{1.68}                       & 87.17\std{1.67}                       & 87.10\std{1.68}                       & 96.28\std{0.92}                      & 96.34\std{0.81}                      \\
                        &                       & \textbf{Medformer}        & \underline{89.62\std{0.81}}           & \underline{89.68\std{0.78}}           & \underline{89.62\std{0.81}}           & \underline{89.62\std{0.81}}           & 96.41\std{0.35}                      & 96.51\std{0.33}                      \\
      \cmidrule{2-9}
                        & \multirow{4}{*}{TSA}  & \textbf{iTransformer-TSA} & 67.50\std{5.50}                       & 65.30\std{5.92}                       & 65.55\std{6.34}                       & 65.07\std{5.92}                       & 72.39\std{6.61}                      & 60.80\std{7.76}                      \\
                        &                       & \textbf{PatchTST-TSA}     & 62.58\std{5.32}                       & 59.34\std{4.26}                       & 59.40\std{3.85}                       & 58.08\std{3.28}                       & 65.90\std{5.71}                      & 51.57\std{3.79}                      \\
                        &                       & \textbf{Reformer-TSA}     & 69.37\std{6.96}                       & 67.28\std{6.35}                       & 63.56\std{1.59}                       & 62.93\std{3.28}                       & 72.01\std{2.61}                      & 62.14\std{1.83}                      \\
                        &                       & \textbf{TimesFM-TSA}      & 84.11\std{10.88}                      & 82.18\std{11.74}                      & 84.75\std{10.99}                      & 82.78\std{11.75}                      & 90.94\std{9.56}                      & 80.71\std{17.82}                     \\
      \cmidrule{2-9}
                        & GA                    & \textbf{\name (Ours)}     & \textbf{88.94\std{8.78}}              & \textbf{89.47\std{7.76}}              & \textbf{86.75\std{12.22}}             & \textbf{87.30\std{11.36}}             & \textbf{95.93\std{4.59}}             & \textbf{93.06\std{8.18}}             \\

      \midrule
      \multirow{16}{*}{\makecell{\textbf{PTB}                                                                                                                                                                                                                                                                             \\ (2-Classes)}} 
                        & \multirow{11}{*}{TSM} & \textbf{Autoformer}       & 73.35\std{2.10}                       & 72.11\std{2.89}                       & 63.24\std{3.17}                       & 63.69\std{3.84}                       & 78.54\std{3.48}                      & 74.25\std{3.53}                      \\
                        &                       & \textbf{Crossformer}      & 80.17\std{3.79}                       & 85.04\std{1.83}                       & 71.25\std{6.29}                       & 72.75\std{7.19}                       & 88.55\std{3.45}                      & 87.31\std{3.25}                      \\
                        &                       & \textbf{FEDformer}        & 76.05\std{2.54}                       & 77.58\std{3.61}                       & 66.10\std{3.55}                       & 67.14\std{4.37}                       & 85.93\std{4.31}                      & 82.59\std{5.42}                      \\
                        &                       & \textbf{Informer}         & 78.69\std{1.68}                       & 82.87\std{1.02}                       & 69.19\std{2.90}                       & 70.84\std{3.47}                       & 92.09\std{0.53}                      & 90.02\std{0.60}                      \\
                        &                       & \textbf{iTransformer}     & 83.89\std{0.71}                       & 88.25\std{1.18}                       & 76.39\std{1.01}                       & 79.06\std{1.06}                       & 91.18\std{1.16}                      & 90.93\std{0.98}                      \\
                        &                       & \textbf{MTST}             & 76.59\std{1.90}                       & 79.88\std{1.90}                       & 66.31\std{2.95}                       & 67.38\std{3.71}                       & 86.86\std{2.75}                      & 83.75\std{2.84}                      \\
                        &                       & \textbf{Nonformer}        & 78.66\std{0.49}                       & 82.77\std{0.86}                       & 69.12\std{0.87}                       & 70.90\std{1.00}                       & 89.37\std{2.51}                      & 86.67\std{2.38}                      \\
                        &                       & \textbf{PatchTST}         & 74.74\std{1.62}                       & 76.94\std{1.51}                       & 63.89\std{2.71}                       & 64.36\std{3.38}                       & 88.79\std{0.91}                      & 83.39\std{0.96}                      \\
                        &                       & \textbf{Reformer}         & 77.96\std{2.13}                       & 81.72\std{1.61}                       & 68.20\std{3.35}                       & 69.65\std{3.88}                       & 91.13\std{0.74}                      & 88.42\std{1.30}                      \\
                        &                       & \textbf{Transformer}      & 77.37\std{1.02}                       & 81.84\std{0.66}                       & 67.14\std{1.80}                       & 68.47\std{2.19}                       & 90.08\std{1.76}                      & 87.22\std{1.68}                      \\
                        &                       & \textbf{Medformer}        & 83.50\std{2.01}                       & 85.19\std{0.94}                       & 77.11\std{3.39}                       & 79.18\std{3.31}                       & 92.81\std{1.48}                      & 90.32\std{1.54}                      \\
      \cmidrule{2-9}
                        & \multirow{4}{*}{TSA}  & \textbf{iTransformer-TSA} & 88.64\std{2.15}                       & 79.41\std{5.58}                       & 75.74\std{7.35}                       & 75.64\std{5.13}                       & 88.97\std{5.31}                      & 97.39\std{2.04}                      \\
                        &                       & \textbf{PatchTST-TSA}     & 73.09\std{16.71}                      & 66.78\std{7.98}                       & 76.45\std{4.32}                       & 62.87\std{12.19}                      & 88.72\std{1.33}                      & 98.36\std{0.20}                      \\
                        &                       & \textbf{Reformer-TSA}     & 88.67\std{1.13}                       & 72.22\std{2.71}                       & 67.04\std{5.11}                       & 68.44\std{3.75}                       & 86.54\std{1.46}                      & 97.82\std{0.39}                      \\
                        &                       & \textbf{TimesFM-TSA}      & 92.17\std{1.88}                       & 83.05\std{6.81}                       & 84.46\std{5.45}                       & 83.63\std{6.02}                       & 91.77\std{3.13}                      & 98.01\std{0.16}                      \\
      \cmidrule{2-9}
                        & GA                    & \textbf{\name (Ours)}     & \underline{\textbf{95.68\std{1.26}}}  & \underline{\textbf{94.24\std{3.65}}}  & \underline{\textbf{90.80\std{1.24}}}  & \underline{\textbf{92.35\std{2.15}}}  & \underline{\textbf{96.54\std{1.82}}} & \underline{\textbf{98.82\std{0.77}}} \\

      \midrule
      \multirow{16}{*}{\makecell{\textbf{PTB-XL}                                                                                                                                                                                                                                                                          \\ (5-Classes)}} 
                        & \multirow{11}{*}{TSM} & \textbf{Autoformer}       & 61.68\std{2.72}                       & 51.60\std{1.64}                       & 49.10\std{1.52}                       & 48.85\std{2.27}                       & 82.04\std{1.44}                      & 51.93\std{1.71}                      \\
                        &                       & \textbf{Crossformer}      & 73.30\std{0.14}                       & 65.06\std{0.35}                       & 61.23\std{0.33}                       & 62.59\std{0.14}                       & 90.02\std{0.06}                      & 67.43\std{0.22}                      \\
                        &                       & \textbf{FEDformer}        & 57.20\std{9.47}                       & 52.38\std{6.09}                       & 49.04\std{7.26}                       & 47.89\std{8.44}                       & 82.13\std{4.17}                      & 52.31\std{7.03}                      \\
                        &                       & \textbf{Informer}         & 71.43\std{0.32}                       & 62.64\std{0.60}                       & 59.12\std{0.47}                       & 60.44\std{0.43}                       & 88.65\std{0.09}                      & 64.76\std{0.17}                      \\
                        &                       & \textbf{iTransformer}     & 69.28\std{0.22}                       & 59.59\std{0.45}                       & 54.62\std{0.18}                       & 56.20\std{0.19}                       & 86.71\std{0.10}                      & 60.27\std{0.21}                      \\
                        &                       & \textbf{MTST}             & 72.14\std{0.27}                       & 63.84\std{0.72}                       & 60.01\std{0.81}                       & 61.43\std{0.38}                       & 88.97\std{0.33}                      & 65.83\std{0.51}                      \\
                        &                       & \textbf{Nonformer}        & 70.56\std{0.55}                       & 61.57\std{0.66}                       & 57.75\std{0.72}                       & 59.10\std{0.66}                       & 88.32\std{0.36}                      & 63.40\std{0.79}                      \\
                        &                       & \textbf{PatchTST}         & 73.23\std{0.25}                       & 65.70\std{0.64}                       & 60.82\std{0.76}                       & 62.61\std{0.34}                       & 89.74\std{0.19}                      & 67.32\std{0.22}                      \\
                        &                       & \textbf{Reformer}         & 71.72\std{0.43}                       & 63.12\std{1.02}                       & 59.20\std{0.75}                       & 60.69\std{0.18}                       & 88.80\std{0.24}                      & 64.72\std{0.47}                      \\
                        &                       & \textbf{Transformer}      & 70.59\std{0.44}                       & 61.57\std{0.65}                       & 57.62\std{0.35}                       & 59.05\std{0.25}                       & 88.21\std{0.16}                      & 63.36\std{0.29}                      \\
                        &                       & \textbf{Medformer}        & 72.87\std{0.23}                       & 64.14\std{0.42}                       & 60.60\std{0.46}                       & 62.02\std{0.37}                       & 89.66\std{0.13}                      & 66.39\std{0.22}                      \\
      \cmidrule{2-9}
                        & \multirow{4}{*}{TSA}  & \textbf{iTransformer-TSA} & 63.85\std{1.38}                       & 55.22\std{2.65}                       & 45.16\std{2.09}                       & 45.96\std{2.11}                       & 83.02\std{0.72}                      & 52.32\std{1.50}                      \\
                        &                       & \textbf{PatchTST-TSA}     & 56.98\std{3.32}                       & 50.71\std{1.43}                       & 45.90\std{1.79}                       & 42.97\std{2.93}                       & 81.21\std{1.03}                      & 50.08\std{1.85}                      \\
                        &                       & \textbf{Reformer-TSA}     & 56.75\std{8.66}                       & 47.31\std{4.27}                       & 38.56\std{5.29}                       & 36.84\std{8.58}                       & 78.79\std{1.09}                      & 45.73\std{2.01}                      \\
                        &                       & \textbf{TimesFM-TSA}      & 71.14\std{0.39}                       & 62.69\std{0.74}                       & 57.75\std{0.49}                       & 59.61\std{0.51}                       & 88.33\std{0.27}                      & 63.06\std{1.16}                      \\
      \cmidrule{2-9}
                        & GA                    & \textbf{\name (Ours)}     & \underline{\textbf{82.73\std{0.71}}}  & \underline{\textbf{78.71\std{0.97}}}  & \underline{\textbf{77.07\std{0.89}}}  & \underline{\textbf{77.83\std{0.92}}}  & \underline{\textbf{94.02\std{0.29}}} & \underline{\textbf{80.86\std{0.74}}} \\
      \bottomrule
    \end{tabular}
  }
  \vskip -0.1in
\end{table*}

%% file: tables/adapting.tex
{
\begin{table}[!ht]
    \centering
    \def\arraystretch{1.0}
    \caption{\textbf{Results on adapting to unseen datasets.}
        Best results are highlighted in \textbf{bold}.
    }
    \label{tab:adaptation}
    \resizebox{\textwidth}{!}{%
        \color{blue}
        \begin{tabular}{cccllllll}
            \toprule
            \textbf{Datasets}                        & \textbf{Model}                               & \textbf{$k$ factor} & \textbf{Accuracy}         & \textbf{Precision}        & \textbf{Recall}           & \textbf{F1 score}         & \textbf{AUROC}           & \textbf{AUPRC}           \\

            \midrule
            \multirow{11}{*}{\textbf{ECG200}}        & \textbf{TimesFM-TSA}                         & N/A                 & 66.32\std{4.71}           & 40.89\std{15.98}          & 52.95\std{11.39}          & 45.23\std{12.63}          & 80.26\std{9.79}          & 91.24\std{5.22}          \\
            \cmidrule{2-9}
                                                     & \multirow{10}{*}{\makecell{\textbf{FORMED}}} & 4                   & 52.63\std{18.23}          & 55.99\std{17.59}          & 56.41\std{17.19}          & 51.28\std{17.23}          & 65.90\std{18.05}         & 81.64\std{10.65}         \\
                                                     &                                              & 8                   & 43.16\std{15.52}          & 43.68\std{11.08}          & 45.00\std{10.89}          & 39.40\std{14.06}          & 45.13\std{15.84}         & 71.92\std{8.59}          \\
                                                     &                                              & 16                  & 57.89\std{11.77}          & 49.65\std{16.85}          & 51.28\std{17.36}          & 49.86\std{16.41}          & 68.46\std{20.1}          & 85.77\std{8.72}          \\
                                                     &                                              & 32                  & 64.21\std{12.57}          & 54.57\std{18.70}          & 55.90\std{17.89}          & 54.46\std{18.12}          & 63.33\std{14.72}         & 80.85\std{8.63}          \\
                                                     &                                              & 64                  & 70.53\std{4.71}           & 65.73\std{5.64}           & 62.31\std{5.45}           & 62.24\std{6.58}           & 73.85\std{7.23}          & 88.05\std{4.42}          \\
                                                     &                                              & 128                 & 70.53\std{4.71}           & 65.39\std{6.96}           & 60.51\std{5.53}           & 60.59\std{6.47}           & 75.90\std{4.19}          & 89.96\std{1.81}          \\
                                                     &                                              & 256                 & 84.21\std{8.32}           & 82.40\std{8.57}           & 82.18\std{14.44}          & 80.56\std{12.31}          & 89.23\std{11.13}         & 95.35\std{5.06}          \\
                                                     &                                              & 512                 & 87.37\std{2.88}           & 85.38\std{2.94}           & 88.08\std{6.01}           & 85.87\std{3.91}           & 95.90\std{4.66}          & 98.27\std{1.97}          \\
                                                     &                                              & 1024                & \textbf{88.42\std{2.35}}  & \textbf{86.67\std{1.86}}  & \textbf{91.54\std{1.72}}  & \textbf{87.65\std{2.33}}  & \textbf{98.21\std{1.15}} & \textbf{99.23\std{0.50}} \\
                                                     &                                              & 2048                & 87.37\std{4.71}           & 85.41\std{4.87}           & 88.97\std{5.31}           & 86.25\std{4.97}           & 95.90\std{3.56}          & 98.22\std{1.56}          \\

            \midrule
            \multirow{11}{*}{\textbf{StandWalkJump}} & \textbf{TimesFM-TSA}                         & N/A                 & 33.33\std{9.43}           & 33.83\std{12.62}          & 33.33\std{9.43}           & 32.41\std{10.64}          & 52.00\std{5.79}          & 44.89\std{4.97}          \\
            \cmidrule{2-9}
                                                     & \multirow{10}{*}{\makecell{\textbf{FORMED}}} & 4                   & 21.33\std{7.30}           & 19.23\std{7.84}           & 21.33\std{7.30}           & 19.69\std{7.22}           & 44.00\std{10.66}         & 42.07\std{7.44}          \\
                                                     &                                              & 8                   & 18.67\std{12.82}          & 19.18\std{13.99}          & 18.67\std{12.82}          & 18.68\std{13.19}          & 46.13\std{12.73}         & 40.63\std{7.32}          \\
                                                     &                                              & 16                  & 36.00\std{10.11}          & 35.22\std{10.23}          & 36.00\std{10.11}          & 34.15\std{8.57}           & 53.60\std{10.18}         & 46.11\std{8.79}          \\
                                                     &                                              & 32                  & 36.00\std{18.01}          & 37.45\std{16.15}          & 36.00\std{18.01}          & 35.87\std{17.37}          & 57.33\std{13.64}         & 53.11\std{15.30}         \\
                                                     &                                              & 64                  & 40.00\std{15.63}          & 36.30\std{15.11}          & 40.00\std{15.63}          & 37.50\std{14.77}          & 56.80\std{14.88}         & 50.30\std{13.12}         \\
                                                     &                                              & 128                 & 46.67\std{10.54}          & 43.96\std{11.27}          & 46.67\std{10.54}          & 44.26\std{12.08}          & 63.60\std{11.72}         & 55.75\std{13.10}         \\
                                                     &                                              & 256                 & 46.67\std{13.33}          & 47.33\std{14.87}          & 46.67\std{13.33}          & 46.29\std{14.25}          & 63.60\std{1.80}          & 53.81\std{2.64}          \\
                                                     &                                              & 512                 & \textbf{64.00\std{11.16}} & \textbf{67.25\std{10.53}} & \textbf{64.00\std{11.16}} & \textbf{64.52\std{10.48}} & \textbf{68.13\std{5.06}} & \textbf{57.27\std{4.25}} \\
                                                     &                                              & 1024                & 54.67\std{7.30}           & 55.24\std{6.28}           & 54.67\std{7.30}           & 54.23\std{7.44}           & 62.13\std{1.66}          & 53.30\std{1.42}          \\
                                                     &                                              & 2048                & 60.00\std{6.67}           & 62.29\std{7.03}           & 60.00\std{6.67}           & 60.12\std{6.63}           & 65.33\std{2.98}          & 56.34\std{2.27}          \\

            \bottomrule
        \end{tabular}
    }
    \vskip -0.1in
\end{table}
}

%% file: sections/100-checklist.tex
\section*{NeurIPS Paper Checklist}

\begin{enumerate}

    \item {\bf Claims}
    \item[] Question: Do the main claims made in the abstract and introduction accurately reflect the paper's contributions and scope?
    \item[] Answer: \answerYes{} 
    \item[] Justification: The abstract and introduction accurately reflect the paper's contributions and scope.
    \item[] Guidelines:
          \begin{itemize}
              \item The answer NA means that the abstract and introduction do not include the claims made in the paper.
              \item The abstract and/or introduction should clearly state the claims made, including the contributions made in the paper and important assumptions and limitations. A No or NA answer to this question will not be perceived well by the reviewers.
              \item The claims made should match theoretical and experimental results, and reflect how much the results can be expected to generalize to other settings.
              \item It is fine to include aspirational goals as motivation as long as it is clear that these goals are not attained by the paper.
          \end{itemize}

    \item {\bf Limitations}
    \item[] Question: Does the paper discuss the limitations of the work performed by the authors?
    \item[] Answer: \answerYes{} 
    \item[] Justification: The limitations are discussed in \cref{sec:conclusion}.
    \item[] Guidelines:
          \begin{itemize}
              \item The answer NA means that the paper has no limitation while the answer No means that the paper has limitations, but those are not discussed in the paper.
              \item The authors are encouraged to create a separate "Limitations" section in their paper.
              \item The paper should point out any strong assumptions and how robust the results are to violations of these assumptions (e.g., independence assumptions, noiseless settings, model well-specification, asymptotic approximations only holding locally). The authors should reflect on how these assumptions might be violated in practice and what the implications would be.
              \item The authors should reflect on the scope of the claims made, e.g., if the approach was only tested on a few datasets or with a few runs. In general, empirical results often depend on implicit assumptions, which should be articulated.
              \item The authors should reflect on the factors that influence the performance of the approach. For example, a facial recognition algorithm may perform poorly when image resolution is low or images are taken in low lighting. Or a speech-to-text system might not be used reliably to provide closed captions for online lectures because it fails to handle technical jargon.
              \item The authors should discuss the computational efficiency of the proposed algorithms and how they scale with dataset size.
              \item If applicable, the authors should discuss possible limitations of their approach to address problems of privacy and fairness.
              \item While the authors might fear that complete honesty about limitations might be used by reviewers as grounds for rejection, a worse outcome might be that reviewers discover limitations that aren't acknowledged in the paper. The authors should use their best judgment and recognize that individual actions in favor of transparency play an important role in developing norms that preserve the integrity of the community. Reviewers will be specifically instructed to not penalize honesty concerning limitations.
          \end{itemize}

    \item {\bf Theory assumptions and proofs}
    \item[] Question: For each theoretical result, does the paper provide the full set of assumptions and a complete (and correct) proof?
    \item[] Answer: \answerNA{} 
    \item[] Justification: There are no theoretical results in the paper.
    \item[] Guidelines:
          \begin{itemize}
              \item The answer NA means that the paper does not include theoretical results.
              \item All the theorems, formulas, and proofs in the paper should be numbered and cross-referenced.
              \item All assumptions should be clearly stated or referenced in the statement of any theorems.
              \item The proofs can either appear in the main paper or the supplemental material, but if they appear in the supplemental material, the authors are encouraged to provide a short proof sketch to provide intuition.
              \item Inversely, any informal proof provided in the core of the paper should be complemented by formal proofs provided in appendix or supplemental material.
              \item Theorems and Lemmas that the proof relies upon should be properly referenced.
          \end{itemize}

    \item {\bf Experimental result reproducibility}
    \item[] Question: Does the paper fully disclose all the information needed to reproduce the main experimental results of the paper to the extent that it affects the main claims and/or conclusions of the paper (regardless of whether the code and data are provided or not)?
    \item[] Answer: \answerYes{} 
    \item[] Justification: All details are described in \cref{sec:experiment} and \cref{sec:experiment-setup}.
    \item[] Guidelines:
          \begin{itemize}
              \item The answer NA means that the paper does not include experiments.
              \item If the paper includes experiments, a No answer to this question will not be perceived well by the reviewers: Making the paper reproducible is important, regardless of whether the code and data are provided or not.
              \item If the contribution is a dataset and/or model, the authors should describe the steps taken to make their results reproducible or verifiable.
              \item Depending on the contribution, reproducibility can be accomplished in various ways. For example, if the contribution is a novel architecture, describing the architecture fully might suffice, or if the contribution is a specific model and empirical evaluation, it may be necessary to either make it possible for others to replicate the model with the same dataset, or provide access to the model. In general. releasing code and data is often one good way to accomplish this, but reproducibility can also be provided via detailed instructions for how to replicate the results, access to a hosted model (e.g., in the case of a large language model), releasing of a model checkpoint, or other means that are appropriate to the research performed.
              \item While NeurIPS does not require releasing code, the conference does require all submissions to provide some reasonable avenue for reproducibility, which may depend on the nature of the contribution. For example
                    \begin{enumerate}
                        \item If the contribution is primarily a new algorithm, the paper should make it clear how to reproduce that algorithm.
                        \item If the contribution is primarily a new model architecture, the paper should describe the architecture clearly and fully.
                        \item If the contribution is a new model (e.g., a large language model), then there should either be a way to access this model for reproducing the results or a way to reproduce the model (e.g., with an open-source dataset or instructions for how to construct the dataset).
                        \item We recognize that reproducibility may be tricky in some cases, in which case authors are welcome to describe the particular way they provide for reproducibility. In the case of closed-source models, it may be that access to the model is limited in some way (e.g., to registered users), but it should be possible for other researchers to have some path to reproducing or verifying the results.
                    \end{enumerate}
          \end{itemize}

    \item {\bf Open access to data and code}
    \item[] Question: Does the paper provide open access to the data and code, with sufficient instructions to faithfully reproduce the main experimental results, as described in supplemental material?
    \item[] Answer: \answerYes{} 
    \item[] Justification: The data can be obtained from the authors of \citet{Medformer} and openly available UCR UEA archive, as described in \cref{sec:experiment-setup}. We will provide public access to code upon acceptance.
    \item[] Guidelines:
          \begin{itemize}
              \item The answer NA means that paper does not include experiments requiring code.
              \item Please see the NeurIPS code and data submission guidelines (\url{https://nips.cc/public/guides/CodeSubmissionPolicy}) for more details.
              \item While we encourage the release of code and data, we understand that this might not be possible, so “No” is an acceptable answer. Papers cannot be rejected simply for not including code, unless this is central to the contribution (e.g., for a new open-source benchmark).
              \item The instructions should contain the exact command and environment needed to run to reproduce the results. See the NeurIPS code and data submission guidelines (\url{https://nips.cc/public/guides/CodeSubmissionPolicy}) for more details.
              \item The authors should provide instructions on data access and preparation, including how to access the raw data, preprocessed data, intermediate data, and generated data, etc.
              \item The authors should provide scripts to reproduce all experimental results for the new proposed method and baselines. If only a subset of experiments are reproducible, they should state which ones are omitted from the script and why.
              \item At submission time, to preserve anonymity, the authors should release anonymized versions (if applicable).
              \item Providing as much information as possible in supplemental material (appended to the paper) is recommended, but including URLs to data and code is permitted.
          \end{itemize}

    \item {\bf Experimental setting/details}
    \item[] Question: Does the paper specify all the training and test details (e.g., data splits, hyperparameters, how they were chosen, type of optimizer, etc.) necessary to understand the results?
    \item[] Answer: \answerYes{} 
    \item[] Justification: All details are described in \cref{sec:experiment} and \cref{sec:experiment-setup}.
    \item[] Guidelines:
          \begin{itemize}
              \item The answer NA means that the paper does not include experiments.
              \item The experimental setting should be presented in the core of the paper to a level of detail that is necessary to appreciate the results and make sense of them.
              \item The full details can be provided either with the code, in appendix, or as supplemental material.
          \end{itemize}

    \item {\bf Experiment statistical significance}
    \item[] Question: Does the paper report error bars suitably and correctly defined or other appropriate information about the statistical significance of the experiments?
    \item[] Answer: \answerYes{} 
    \item[] Justification: The error bars or raw data points are plotted in the figures (\cref{fig:comparison-f1,fig:comparison-accuracy,fig:comparison-precision,fig:comparison-recall,fig:comparison-auroc,fig:comparison-auprc,fig:adapting}) and the standard deviation is reported in the tables (\cref{tab:full-result,tab:adaptation}).
    \item[] Guidelines:
          \begin{itemize}
              \item The answer NA means that the paper does not include experiments.
              \item The authors should answer "Yes" if the results are accompanied by error bars, confidence intervals, or statistical significance tests, at least for the experiments that support the main claims of the paper.
              \item The factors of variability that the error bars are capturing should be clearly stated (for example, train/test split, initialization, random drawing of some parameter, or overall run with given experimental conditions).
              \item The method for calculating the error bars should be explained (closed form formula, call to a library function, bootstrap, etc.)
              \item The assumptions made should be given (e.g., Normally distributed errors).
              \item It should be clear whether the error bar is the standard deviation or the standard error of the mean.
              \item It is OK to report 1-sigma error bars, but one should state it. The authors should preferably report a 2-sigma error bar than state that they have a 96\% CI, if the hypothesis of Normality of errors is not verified.
              \item For asymmetric distributions, the authors should be careful not to show in tables or figures symmetric error bars that would yield results that are out of range (e.g. negative error rates).
              \item If error bars are reported in tables or plots, The authors should explain in the text how they were calculated and reference the corresponding figures or tables in the text.
          \end{itemize}

    \item {\bf Experiments compute resources}
    \item[] Question: For each experiment, does the paper provide sufficient information on the computer resources (type of compute workers, memory, time of execution) needed to reproduce the experiments?
    \item[] Answer: \answerYes{} 
    \item[] Justification: We provide the environment used to run the experiments in \cref{sec:experiment-setup}. The exact training time is missing due to the use of multiple different machines and there is no consistent way to measure the time of execution.
    \item[] Guidelines:
          \begin{itemize}
              \item The answer NA means that the paper does not include experiments.
              \item The paper should indicate the type of compute workers CPU or GPU, internal cluster, or cloud provider, including relevant memory and storage.
              \item The paper should provide the amount of compute required for each of the individual experimental runs as well as estimate the total compute.
              \item The paper should disclose whether the full research project required more compute than the experiments reported in the paper (e.g., preliminary or failed experiments that didn't make it into the paper).
          \end{itemize}

    \item {\bf Code of ethics}
    \item[] Question: Does the research conducted in the paper conform, in every respect, with the NeurIPS Code of Ethics \url{https://neurips.cc/public/EthicsGuidelines}?
    \item[] Answer: \answerYes{} 
    \item[] Justification: We use publicly available and deidentified datasets (as described in \cref{sec:dataset-detail}) in our experiments, and respect the NeurIPS Code of Ethics.
    \item[] Guidelines:
          \begin{itemize}
              \item The answer NA means that the authors have not reviewed the NeurIPS Code of Ethics.
              \item If the authors answer No, they should explain the special circumstances that require a deviation from the Code of Ethics.
              \item The authors should make sure to preserve anonymity (e.g., if there is a special consideration due to laws or regulations in their jurisdiction).
          \end{itemize}

    \item {\bf Broader impacts}
    \item[] Question: Does the paper discuss both potential positive societal impacts and negative societal impacts of the work performed?
    \item[] Answer: \answerNo{} 
    \item[] Justification: The paper is focused on technical contributions and propose a solution to tackle the unique challenges in medical time series classification. Although the application field is healthcare, the paper does not directly relate to any specific application. The authors do not foresee any negative societal impacts of the work performed.
    \item[] Guidelines:
          \begin{itemize}
              \item The answer NA means that there is no societal impact of the work performed.
              \item If the authors answer NA or No, they should explain why their work has no societal impact or why the paper does not address societal impact.
              \item Examples of negative societal impacts include potential malicious or unintended uses (e.g., disinformation, generating fake profiles, surveillance), fairness considerations (e.g., deployment of technologies that could make decisions that unfairly impact specific groups), privacy considerations, and security considerations.
              \item The conference expects that many papers will be foundational research and not tied to particular applications, let alone deployments. However, if there is a direct path to any negative applications, the authors should point it out. For example, it is legitimate to point out that an improvement in the quality of generative models could be used to generate deepfakes for disinformation. On the other hand, it is not needed to point out that a generic algorithm for optimizing neural networks could enable people to train models that generate Deepfakes faster.
              \item The authors should consider possible harms that could arise when the technology is being used as intended and functioning correctly, harms that could arise when the technology is being used as intended but gives incorrect results, and harms following from (intentional or unintentional) misuse of the technology.
              \item If there are negative societal impacts, the authors could also discuss possible mitigation strategies (e.g., gated release of models, providing defenses in addition to attacks, mechanisms for monitoring misuse, mechanisms to monitor how a system learns from feedback over time, improving the efficiency and accessibility of ML).
          \end{itemize}

    \item {\bf Safeguards}
    \item[] Question: Does the paper describe safeguards that have been put in place for responsible release of data or models that have a high risk for misuse (e.g., pretrained language models, image generators, or scraped datasets)?
    \item[] Answer: \answerNA{} 
    \item[] Justification: The paper does not fall into the category of high-risk models.
    \item[] Guidelines:
          \begin{itemize}
              \item The answer NA means that the paper poses no such risks.
              \item Released models that have a high risk for misuse or dual-use should be released with necessary safeguards to allow for controlled use of the model, for example by requiring that users adhere to usage guidelines or restrictions to access the model or implementing safety filters.
              \item Datasets that have been scraped from the Internet could pose safety risks. The authors should describe how they avoided releasing unsafe images.
              \item We recognize that providing effective safeguards is challenging, and many papers do not require this, but we encourage authors to take this into account and make a best faith effort.
          \end{itemize}

    \item {\bf Licenses for existing assets}
    \item[] Question: Are the creators or original owners of assets (e.g., code, data, models), used in the paper, properly credited and are the license and terms of use explicitly mentioned and properly respected?
    \item[] Answer: \answerYes{} 
    \item[] Justification: The paper properly credits the creators of the datasets and models used. The license and terms of use are also not present in the paper, but the authors will provide them along with the code release.
    \item[] Guidelines:
          \begin{itemize}
              \item The answer NA means that the paper does not use existing assets.
              \item The authors should cite the original paper that produced the code package or dataset.
              \item The authors should state which version of the asset is used and, if possible, include a URL.
              \item The name of the license (e.g., CC-BY 4.0) should be included for each asset.
              \item For scraped data from a particular source (e.g., website), the copyright and terms of service of that source should be provided.
              \item If assets are released, the license, copyright information, and terms of use in the package should be provided. For popular datasets, \url{paperswithcode.com/datasets} has curated licenses for some datasets. Their licensing guide can help determine the license of a dataset.
              \item For existing datasets that are re-packaged, both the original license and the license of the derived asset (if it has changed) should be provided.
              \item If this information is not available online, the authors are encouraged to reach out to the asset's creators.
          \end{itemize}

    \item {\bf New assets}
    \item[] Question: Are new assets introduced in the paper well documented and is the documentation provided alongside the assets?
    \item[] Answer: \answerNA{} 
    \item[] Justification: The paper does not release new assets.
    \item[] Guidelines:
          \begin{itemize}
              \item The answer NA means that the paper does not release new assets.
              \item Researchers should communicate the details of the dataset/code/model as part of their submissions via structured templates. This includes details about training, license, limitations, etc.
              \item The paper should discuss whether and how consent was obtained from people whose asset is used.
              \item At submission time, remember to anonymize your assets (if applicable). You can either create an anonymized URL or include an anonymized zip file.
          \end{itemize}

    \item {\bf Crowdsourcing and research with human subjects}
    \item[] Question: For crowdsourcing experiments and research with human subjects, does the paper include the full text of instructions given to participants and screenshots, if applicable, as well as details about compensation (if any)?
    \item[] Answer: \answerNA{} 
    \item[] Justification: The paper does not involve crowdsourcing nor research with human subjects.
    \item[] Guidelines:
          \begin{itemize}
              \item The answer NA means that the paper does not involve crowdsourcing nor research with human subjects.
              \item Including this information in the supplemental material is fine, but if the main contribution of the paper involves human subjects, then as much detail as possible should be included in the main paper.
              \item According to the NeurIPS Code of Ethics, workers involved in data collection, curation, or other labor should be paid at least the minimum wage in the country of the data collector.
          \end{itemize}

    \item {\bf Institutional review board (IRB) approvals or equivalent for research with human subjects}
    \item[] Question: Does the paper describe potential risks incurred by study participants, whether such risks were disclosed to the subjects, and whether Institutional Review Board (IRB) approvals (or an equivalent approval/review based on the requirements of your country or institution) were obtained?
    \item[] Answer: \answerNA{} 
    \item[] Justification: The paper does not involve crowdsourcing nor research with human subjects.
    \item[] Guidelines:
          \begin{itemize}
              \item The answer NA means that the paper does not involve crowdsourcing nor research with human subjects.
              \item Depending on the country in which research is conducted, IRB approval (or equivalent) may be required for any human subjects research. If you obtained IRB approval, you should clearly state this in the paper.
              \item We recognize that the procedures for this may vary significantly between institutions and locations, and we expect authors to adhere to the NeurIPS Code of Ethics and the guidelines for their institution.
              \item For initial submissions, do not include any information that would break anonymity (if applicable), such as the institution conducting the review.
          \end{itemize}

    \item {\bf Declaration of LLM usage}
    \item[] Question: Does the paper describe the usage of LLMs if it is an important, original, or non-standard component of the core methods in this research? Note that if the LLM is used only for writing, editing, or formatting purposes and does not impact the core methodology, scientific rigorousness, or originality of the research, declaration is not required.
    \item[] Answer: \answerNA{} 
    \item[] Justification: The LLMs are not used for the core methods in this research.
    \item[] Guidelines:
          \begin{itemize}
              \item The answer NA means that the core method development in this research does not involve LLMs as any important, original, or non-standard components.
              \item Please refer to our LLM policy (\url{https://neurips.cc/Conferences/2025/LLM}) for what should or should not be described.
          \end{itemize}

\end{enumerate}